\documentclass[10pt,twocolumn,letterpaper]{article}
\pdfoutput=1
\usepackage{iccv}
\usepackage{times}
\usepackage{epsfig}
\usepackage{graphicx}
\usepackage{amsmath}
\usepackage{amssymb}

\usepackage{amsmath,amsfonts,bm}

\def\1{\bm{1}}

\def\vx{{\bm{x}}}

\DeclareMathAlphabet{\mathsfit}{\encodingdefault}{\sfdefault}{m}{sl}
\SetMathAlphabet{\mathsfit}{bold}{\encodingdefault}{\sfdefault}{bx}{n}

\usepackage{xspace}
\usepackage{xcolor}
\usepackage{enumitem}
\usepackage{bbm}
\usepackage{algorithm}
\usepackage{algorithmic}
\usepackage{listings}
\usepackage{float}
\usepackage{multirow}
\usepackage{color}
\usepackage{colortbl}
\usepackage{xcolor}
\usepackage{pifont}
\usepackage{array}
\usepackage{booktabs}
\usepackage[accsupp]{axessibility}
\usepackage{tocloft}
\usepackage[citecolor=citecolor,pagebackref=true,breaklinks=true,colorlinks,bookmarks=false]{hyperref}
\newcommand\mypara[1]{\vspace{1.0mm}\noindent\textbf{#1}}
\newcommand{\ourmethod}{{\sc {ITI-Gen}}\xspace}
\newcommand{\ourmethodbold}{{\textbf{\textsc {ITI-Gen}}}\xspace}

\newcommand{\append}[1]{{\color{red} #1}\xspace}
\newcommand{\s}[1]{{\color{red}#1}\xspace}
\newcommand{\cmark}{\ding{51}}%
\newcommand{\xmark}{\ding{55}}%

\newcommand{\nocontentsline}[3]{}
\newcommand{\tocless}[2]{\bgroup\let\addcontentsline=\nocontentsline#1{#2}\egroup}
\def\etal{{\em et al.}}

\definecolor{ForestGreen}{rgb}{0.13, 0.55, 0.13}
\definecolor{Maroon}{rgb}{0.69, 0.19, 0.0}
\definecolor{my_pink}{rgb}{1,0.43,0.41}
\definecolor{my_gray}{rgb}{0,0,0.2}
\definecolor{citecolor}{RGB}{30,130,255}
\definecolor{LightCyan}{rgb}{0.88,1,1}

\iccvfinalcopy

\begin{document}
\title{\ourmethodbold: Inclusive Text-to-Image Generation}
\author{
Cheng Zhang $^{1}$ \quad
Xuanbai Chen $^{1}$ \quad
Siqi Chai $^{1}$ \quad
Chen Henry Wu $^{1}$ \quad 
Dmitry Lagun $^{2}$\\ [2pt]
Thabo Beeler $^{2}$ \quad
Fernando De la Torre $^{1}$\\ [6pt]
{$^{1}$ {Carnegie Mellon University} \qquad
$^{2}$ {Google}
} 
}

\maketitle
\begin{abstract}

Text-to-image generative models often reflect the biases of the training data, leading to unequal representations of underrepresented groups. This study investigates \textbf{inclusive} text-to-image generative models that generate images based on human-written prompts and ensure the resulting images are \textbf{uniformly distributed} across attributes of interest. Unfortunately, directly expressing the desired attributes in the prompt often leads to sub-optimal results due to linguistic ambiguity or model misrepresentation. 
Hence, this paper proposes a drastically different approach that adheres to the maxim that {\em ``a picture is worth a thousand words''}. We show that, for some attributes, images can represent concepts more expressively than text. For instance, categories of skin tones are typically hard to specify by text but can be easily represented by example images. Building upon these insights, we propose a novel approach, \textit{\ourmethodbold}\footnote{Project page: \url{https://czhang0528.github.io/iti-gen}}, that leverages readily available reference images for \textbf{I}nclusive \textbf{T}ext-to-\textbf{I}mage \textbf{GEN}eration. The key idea is learning a set of prompt embeddings to generate images that can effectively represent all desired attribute categories. More importantly, \ourmethod requires no model fine-tuning, making it computationally efficient to augment existing text-to-image models. Extensive experiments demonstrate that \ourmethod largely improves over state-of-the-art models to generate inclusive images from a prompt.

\end{abstract}


\tocless 
\section{Introduction}
\label{s_introduction}

\begin{figure}[t]
    \centerline{\includegraphics[width=1\linewidth]{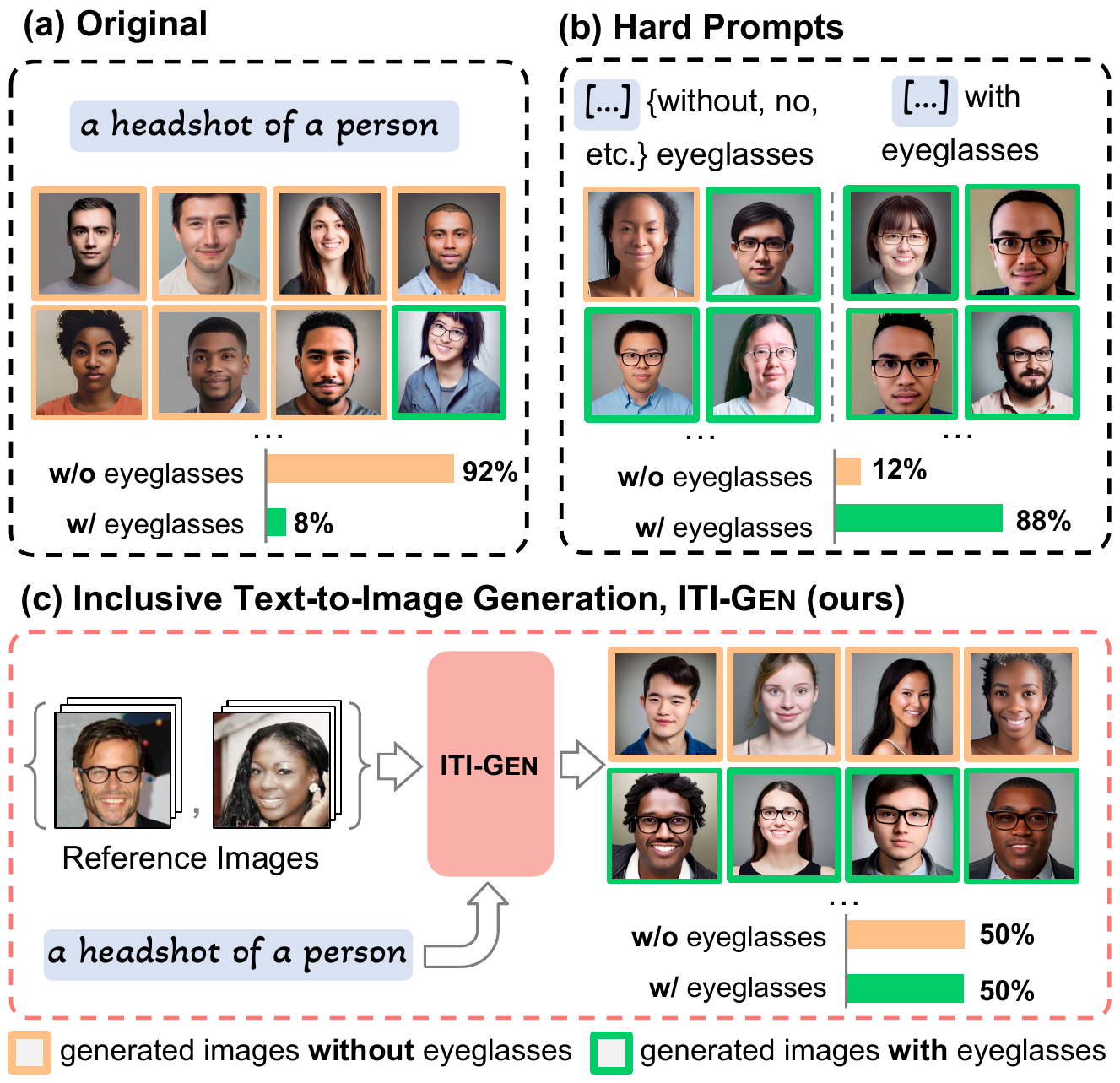}}
    \caption{\small \textbf{(a)} Given a human-written prompt (``\emph{a headshot of a person}''), existing text-to-image models~\cite{rombach2022high} can hardly synthesize pictures representing minority groups (\ie, people with eyeglasses in this example). \textbf{(b)} Conventional hard prompt searching~\cite{ding2021cogview} is sub-optimal due to linguistic ambiguity.
    \textbf{(c)} We address these problems by leveraging a small set of reference images for inclusive text-to-image generation (\ourmethod).
    }
    \vspace{-2mm}
    \label{fig: motivation}
\end{figure}

In recent years we have witnessed a remarkable leap in text-based visual content creation, driven by breakthroughs in generative modeling~\cite{sohl2015deep,ho2020denoising,ramesh2021zero,ramesh2022hierarchical,rombach2022high} and the access to large-scale multimodal datasets~\cite{schuhmann2022laion,karras2019style}. Particularly, publicly released models, such as Stable Diffusion~\cite{rombach2022high}, have matured to the point where they can produce highly realistic images based on human-written prompts.

However, one major drawback of existing text-to-image models is that they inherit biases from the training data~\cite{bianchi2023easily,ramesh2022hierarchical,rombach2022high,cho2022dall,bansal2022well} and thus have yet to exhibit \emph{inclusiveness} --- the generated images based on the input text may reflect stereotypes, leading to the exclusion of certain attributes or minority groups. For instance, given the prompt ``a headshot of a person'', Figure~\ref{fig: motivation}(a) shows how a state-of-the-art system generates about 92$\%$ images of subjects without eyeglasses, and only 8$\%$ with eyeglasses, showing a clear bias towards people without eyeglasses. Alternatively, as shown in Figure~\ref{fig: motivation}(b), one could specify the attribute in the prompt, resulting in better outcomes; however, this will still result in a sub-optimal solution due to linguistic ambiguity. While inclusiveness has been critical to responsible AI, existing text-to-image models are still lagging~\cite{cho2022dall,bansal2022well,petsiuk2022human,parraga2022debiasing,luccioni2023stable}. In this work, we propose a new method that achieves inclusiveness\footnote{Few works~\cite{cho2022dall,bansal2022well} have studied fairness issues in text-to-image generation but mainly focused on social biases (\eg, perceived gender, ethnicity). This paper incorporates a broader spectrum of attributes.} in text-to-image generation using only a few example images, as illustrated in Figure~\ref{fig: motivation}(c).

To advance inclusive generation, a straightforward way is to retrain or fine-tune the model upon request, using \emph{truly} inclusive training data~\cite{dhariwal2021diffusion,zhao2020leveraging}. 
Doing so, however, is insurmountably challenging as collecting large-scale training data that is balanced/inclusive across all attributes of interest is impractical, and training generative models is highly compute-intensive~\cite{schuhmann2022laion,russakovsky2015imagenet,dhariwal2021diffusion}. Another principled approach towards inclusiveness is to specify or enumerate each category in natural language (\ie, hard prompt searching)~\cite{ding2021cogview, petsiuk2022human}.
However, many categories are difficult to specify with natural language (\eg, skin tone) or cannot be well synthesized by the existing models due to linguistic ambiguity or model misrepresentation~\cite{hutchinson2022underspecification}.

At first glance, these seem to paint a grim picture for inclusive text-to-image generation. However, we argue that instead of specifying attributes explicitly using descriptive natural language, images can represent specific concepts or attributes more efficiently. Observing the availability of a shared vision-language embedding in many multimodal generative models~\cite{radford2021learning}, we raise the question: \emph{can we learn inclusive prompt embeddings using images as guidance?}

To achieve this goal, we introduce \ourmethodbold, a novel and practical framework that creates discriminative prompts based on readily available reference images for \textbf{I}nclusive \textbf{T}ext-to-\textbf{I}mage \textbf{GEN}eration. 
Concretely, we leverage the vision-language pre-trained CLIP model~\cite{radford2021learning} to obtain the embeddings of the reference images and learnable prompts. In the joint embedding space, we design a new training objective to align the directions of the image and prompt features. 
The core idea is to translate the visual attribute differences into natural language differences 
such that the generated images based on the learned prompts can effectively represent all desired categories. By equalizing the sampling process over the learned prompts, our method guarantees inclusiveness for text-to-image generation.

We validate our framework with Stable Diffusion~\cite{rombach2022high}. \ourmethod can leverage reference images from different domains, including human faces~\cite{liu2015faceattributes,karkkainen2019fairface,Feng:TRUST:ECCV2022} and scenes~\cite{skorokhodov2021aligning}, to achieve inclusive generation in single or multiple attributes of interest. 
\ourmethod needs neither prompt specification nor model fine-tuning, bypassing the problems of linguistic ambiguity as well as computational complexity. 
Moreover, \ourmethod is compatible with the existing text-based image generation models (\eg, ControlNet~\cite{zhang2023adding} and instruction-based image editing models~\cite{brooks2023instructpix2pix}) in a plug-and-play manner.
To the best of our knowledge, this is the first method that allows inclusive text-to-image generation over a frozen model and obtains competitive results throughout.
\tocless
\section{Related Work}
\label{s_related}

\mypara{Text-to-Image Generative Models.}
Text-based image generation has been widely studied with numerous model architectures and learning paradigms~\cite{mansimov2015generating,reed2016generative,tao2022df,ramesh2021zero,gafni2022make,yu2022scaling,ding2021cogview,ding2022cogview2,chang2023muse,sohl2015deep,yang2022diffusion,croitoru2022diffusion,dhariwal2021diffusion,kingma2021variational}. Recently, the overwhelming success of diffusion-based text-to-image models~\cite{ramesh2022hierarchical,saharia2022photorealistic,ramesh2022hierarchical,nichol2021glide} has attracted significant attention. A key factor to this success is their ability to deal with large-scale multimodal datasets~\cite{schuhmann2022laion,karras2019style,chen2015microsoft}.
Thus, questions concerning inclusiveness while learning with biased datasets remain a crucial open problem~\cite{cho2022dall,bansal2022well,agarwal2021evaluating}. 

\begin{figure*}[t]
    \centerline{\includegraphics[width=1\linewidth]{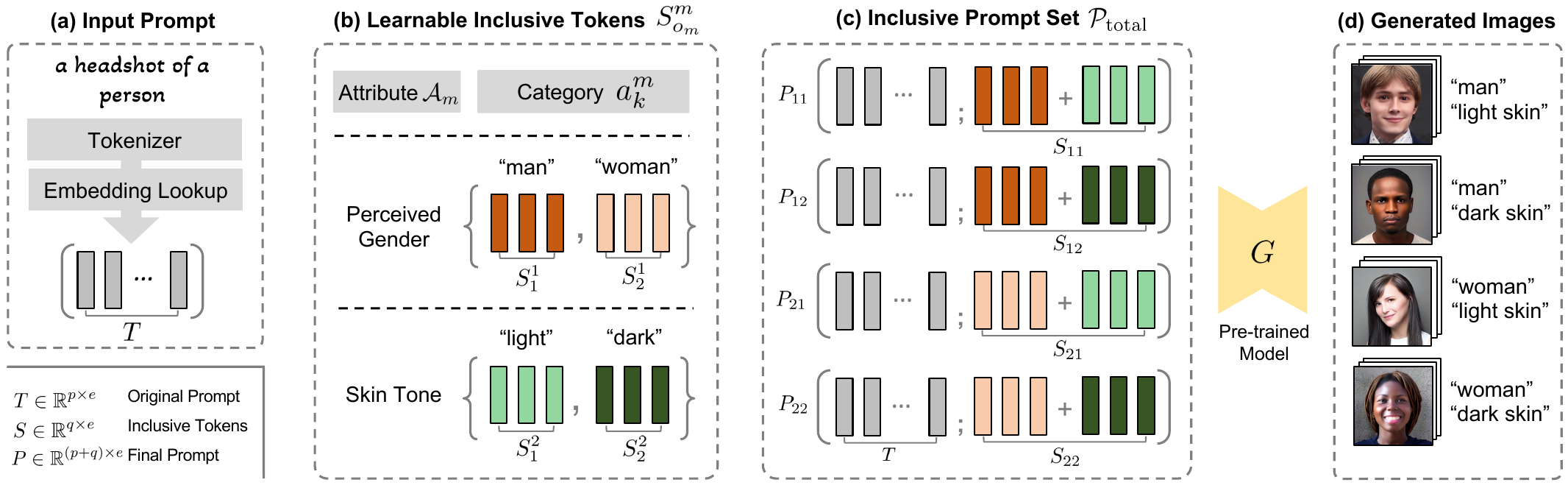}}
    \caption{\small \textbf{Illustration of Inclusive Text-to-Image GENeration (\ourmethodbold)} with the example of two binary attributes: \emph{perceived gender} and \emph{skin tone}. \textbf{(a)} Given an input prompt, \textbf{(b)} \ourmethod learns discriminative token embeddings to represent each category of every target attribute. \textbf{(c)} By injecting the learned tokens after the original input prompt, \ourmethod synthesizes an inclusive prompt set that can be used to \textbf{(d)} sample equal (or controllable) numbers of images for any category combination. Further, our framework can be easily extended to multi-category multi-attribute scenarios of inclusive text-to-image generation. Note that, in practice, multi-category skin tones beyond $\{\text{``light''},\text{``dark''} \}$ as in this example may be challenging to specify with language (see Figure~\ref{fig:approach}). Please see 
    Section~\s{3.1}
    for details.
    }
    \label{fig:overview}
    \vspace{-2mm}
\end{figure*}

\mypara{Bias Mitigation in Text-to-Image Generation.}
While fairness has been studied extensively in discriminative models ~\cite{wang2020mitigating,wang2019balanced,wang2020towards,liang2022advances}, research on developing fair generative models is limited~\cite{zhao2018bias,jain2020imperfect,friedrich2023fair,chuang2023debiasing,luccioni2023stable}. Most efforts focus on GAN-based models~\cite{choi2020fair, ramaswamy2021fair,jalal2021fairness,rangwani2022improving,yu2020inclusive,kenfack2022repfair,xu2018fairgan,tan2020improving,karakas2022fairstyle,maluleke2022studying}, restricting their applicability to the emerging diffusion-based text-to-image models.
Recently, there have been some efforts to address this limitation. For instance, Bansal~\etal~\cite{bansal2022well} proposed to diversify model outputs by ethical intervention\footnote{\eg, appending ``irrespective of their gender'' to the end of a neutral prompt ``a photo of a lawyer'' for generating diverse pictures w.r.t. genders.}. Ding~\etal~\cite{ding2021cogview} proposed to directly add attribute words to the prompt. However, these hard prompt searching methods have limitations such as being opaque and laborious~\cite{bansal2022well}, and not always generating diverse images reliably~\cite{hutchinson2022underspecification,bansal2022well}.
In this work, we incorporate a broad spectrum of attributes beyond social groups. Moreover, we learn inclusive prompts in the continuous embedding space, requiring no hard prompt specification.

To learn a fair generative model, Wu \etal~\cite{wu2022generative} employed off-the-shelf models, such as CLIP~\cite{radford2021learning} and pre-trained classifiers, as guidance. Choi \etal~\cite{choi2020fair} used a reference dataset to train the model via sample re-weighting. In contrast, we use reference data in a drastically different way --- treating the images as proxy signals to guide prompt learning but without retraining the text-to-image model.

\mypara{Image-Guided Prompt Tuning.}
Our method is inspired by Prompt Tuning (PT)~\cite{lester2021power,jia2022visual}. Typically, PT methods insert small learnable modules (\eg, tokens) into the pre-trained models and fine-tune these modules with downstream tasks while freezing the model parameters. Recently, PT has been leveraged in personalized text-to-image generation~\cite{gal2022image,ruiz2022dreambooth,kumari2022multi}. By providing several reference images with the customized subject, they use a special token to represent the object by optimizing the token embedding~\cite{gal2022image,kumari2022multi} or the diffusion models~\cite{ruiz2022dreambooth,kumari2022multi}. This motivates us to learn the specific token embedding for each attribute category for inclusiveness. 
However, we note that the previously mentioned methods for personalization do not effectively capture the attributes in the images. Thus, we propose to optimize the directions of the attribute-specific prompts in the joint vision-language embedding space, bypassing training text-to-image generative models.
\vspace{3mm}
\tocless
\section{Inclusive Text-to-Image Generation}
\label{s_approach}

To drive the progress of Inclusive Text-to-Image Generation, we propose \ourmethod, which creates inclusive prompts that represent various attributes and their combinations. This is particularly challenging for attributes that are difficult to describe in language or underrepresented. To address this, \ourmethod uses readily available reference images as guidance, enabling unambiguous specification of different attributes. Figure~\ref{fig:overview} illustrates the overall framework. In this section, we first introduce the framework of \ourmethod in Section~\s{3.1}, then describe the details of the learning strategy in Section~\s{3.2}, and finally discuss the key properties of \ourmethod in Section~\s{3.3}.

\vspace{3mm}
\tocless
\subsection{Overview}
\label{ss_problemsetup}
\mypara{Problem Statement.}
Given a pre-trained text-to-image generative model $G$ and a human-written prompt (\eg, \emph{``a headshot of a person''}) tokenized as $\bm{T} \in \mathbb{R}^{p \times e}$, where $p$ is the number of tokens and $e$ is the dimension of the embedding space, we aim to sample equal (or controllable) numbers of images that can represent any \emph{category} combination given the \emph{attribute} set $\bm{A}$. Formally,
\begin{align} \label{eq:attribute}
    \bm{A} = \{ \mathcal{A}_m | 1 \leq m \leq M \}; \mathcal{A}_m = \{a^{m}_{k} | 1 \leq k \leq K_m \} 
\end{align}
contains $M$ different attributes (\eg, perceived gender, skin tone, \etc.), where $a_{k}^{m}$ records a mutually exclusive category (\eg, a specific type of skin tone) in attribute $\mathcal{A}_m$ and $K_m$ denotes the number of categories in $\mathcal{A}_m$. Note that $K_m$ may vary among different attributes.

\mypara{Inclusive Prompt Set.} Inspired by \cite{lester2021power,jia2022visual}, we propose prompt tuning for inclusive generation. Specifically, for a given category $a^{m}_{k}$ within attribute $\mathcal{A}_{m}$, we inject $q$ \emph{learnable} tokens $\bm{S}^{m}_{k} \in \mathbb{R}^{q \times e}$ after the original $\bm{T}$ to construct a new prompt $\bm{P}_{k}^{m} = [\bm{T}; \bm{S}^{m}_{k}] \in \mathbb{R}^{(p+q) \times e}$. By querying the model $G$ with $\bm{P}_{k}^{m}$, we can generate images exhibiting the characteristics of the corresponding category $a^{m}_{k}$. To differentiate the new tokens $\bm{S}^{m}_k$ from the original prompt $\bm{T}$, we refer to them as \emph{inclusive tokens}.

When jointly considering $M$ attributes, we aggregate $M$ separate inclusive tokens $\bm{S}^{1}_{o_1}, \bm{S}^{2}_{o_2}, \dots, \bm{S}^{M}_{o_M}$ to represent a specific category combination $(a^1_{o_1}, a^2_{o_2}, \dots, a^M_{o_M})$, \eg, the concept of (``woman'', ``dark skin'', $\dots$, ``young''). We thus expect to create a unique $\bm{S}_{o_1 o_2 \dots o_M}$,
\begin{align} \label{eq:aggregation}
    \bm{S}_{o_1 o_2 \dots o_M} = f (\bm{S}^{1}_{o_1}, \bm{S}^{2}_{o_2}, \dots, \bm{S}^{M}_{o_M})
\end{align}
that can be injected after $\bm{T}$ to generate images for this particular category combination. The aggregation function $f$ in Equation~\ref{eq:aggregation} should be able to take various numbers of attributes while maintaining the permutation invariant property\footnote{That is, the output of $f$ should be the same even if we permute the indices $m$ of the attributes in $\bm{A}$ (cf. Equation~\ref{eq:attribute}).} with respect to attributes. Common options include element-wise average, sum, and max operations. Following~\cite{mikolov2013efficient}, we adopt element-wise sum to preserve the text semantics without losing information\footnote{Please see Appendix~\ref{app_ss_aggregation} for more analysis and other options for aggregating multiple tokens, \eg, concatenation.}.
Finally, we define the \emph{inclusive prompt set} as follows:\vspace{-2mm}
\begin{align} \label{eq:set} 
    \mathcal{P}_\text{total} = \{ &\bm{P}_{o_1 o_2 \dots o_M} = [{\color{blue}{\bm{T}}};\sum_{m=1}^{M} {\color{my_pink}{\bm{S}^{m}_{o_m}}}] \in \mathbb{R}^{(p+q) \times e} ~| \nonumber \\
    & 1 \leq o_1 \leq K_1, \dots, 1 \leq o_M \leq K_M \}.
\end{align}
By uniformly sampling the prompts from $\mathcal{P}_\text{total}$ as the conditions to generate images using the generative model $G$, we achieve inclusiveness across all attributes (see Figure~\ref{fig:overview}). 
\emph{More generally speaking, the distribution of the generated data is directly correlated to the distribution of the prompts, which can be easily controlled.}

In contrast to specifying the category name in discrete language space~\cite{bansal2022well,ding2021cogview}, we optimize prompts entirely in the \emph{continuous} embedding space. Additionally, we only update the attribute-specific embeddings --- the colors {\color{blue}{$\bullet$}} and {\color{my_pink}{$\bullet$}} in Equation~\ref{eq:set} indicate {\color{blue}{frozen}} and {\color{my_pink}{learnable}} parameters, respectively. This decoupled optimization mechanism thus provides the advantage of using the learned inclusive tokens in a plug-and-play manner across various applications, as will be demonstrated in 
Section~\s{3.3} and Section~\s{4.3}. We elaborate on the learning process in the following section.

\vspace{2mm}
\tocless
\subsection{Learning Inclusive Prompts} \label{ss_objective}
\mypara{Reference Image Set.}
We propose using reference images to guide prompt learning, as they can provide more expressive signals to describe attributes that may be challenging to articulate through language. Specifically, we assume the availability of a reference image set $\mathcal{D}^{m}_\text{ref} = \{ (\vx_n^m, y_n^m)\}_{n=1}^{N_m}$ for a target attribute $\mathcal{A}_m$, where $N_m$ is the dataset size and $y_n^m \in \mathcal{A}_m$ (defined in Equation~\ref{eq:attribute}) indicates the category to which $\vx_n$ belongs. When considering multiple attributes, we only need a reference dataset for each attribute, rather than one large balanced dataset with all attribute labels. \emph{This property is extremely beneficial, as it is much easier to obtain a dataset that captures only the distribution of one attribute (i.e., the marginal distribution) rather than one that captures the {joint} distribution of all attributes.}

\mypara{Aligning Prompts to Images with CLIP.}
Given reference image sets for the target attributes, can we learn prompts that align the attributes in the images? Recently, pre-trained large-scale multimodal models have demonstrated strong capabilities in connecting vision and language. One such model is CLIP~\cite{radford2021learning}, which aligns visual concepts with text embeddings by jointly training a text encoder $E_\text{text}$ and an image encoder $E_\text{img}$. The output of the pre-trained CLIP text encoder has also been used as the condition for text-guided image generation~\cite{rombach2022high,ramesh2022hierarchical}, opening up an opportunity to align prompts to reference images without the need to modify the text-to-image models.

One straightforward solution is to maximize the similarity between the prompt and the reference image embeddings in the CLIP space, as suggested by~\cite{radford2021learning}. However, we found it deficient for two reasons. First, this objective forces the prompt to focus on the overall visual information in the images, rather than the specific attribute of interest. Second, the generated images from the learned prompt often exhibit adversarial effects or significant quality degradation, potentially due to image features distorting the prompt embedding. To address these, we propose direction alignment and semantic consistency losses, as described below.

\begin{figure}[t]
    \centerline{\includegraphics[width=1\linewidth]{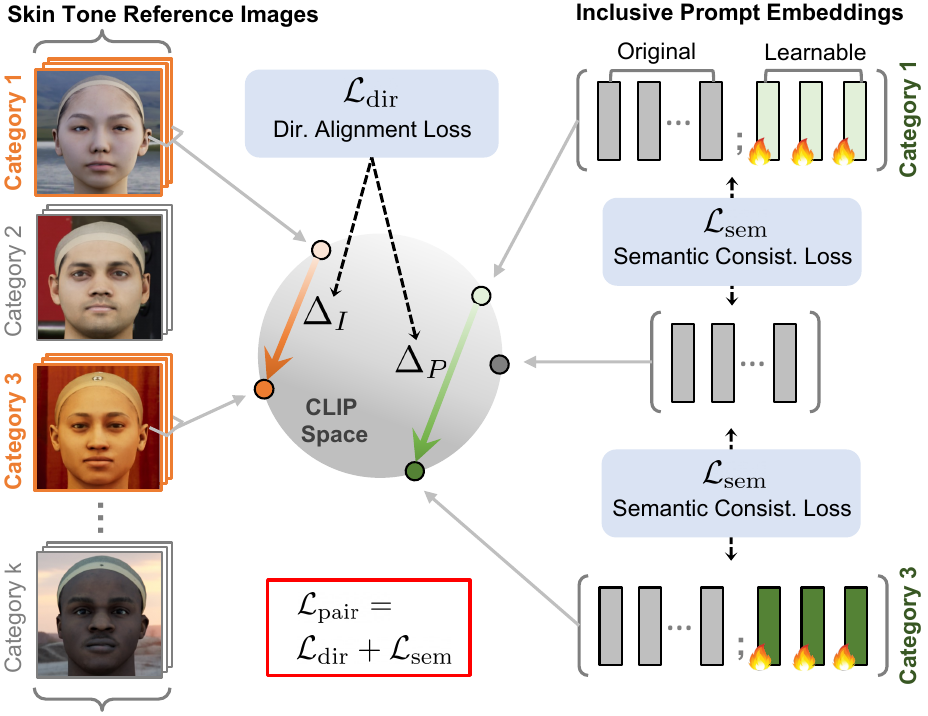}}
    \caption{\small \textbf{Translating visual differences into text embedding differences.} Given reference images of a multi-category attribute (\eg, skin tone), we learn the inclusive tokens by direction alignment between {\color{orange}{\textbf{images}}} and {\color{ForestGreen}{\textbf{prompts}}}, ensuring that the visual difference matches the learned language description. In addition, we propose semantic consistency loss to address language drift. Images are from FAIR benchmark~\cite{Feng:TRUST:ECCV2022}. Details are in 
    Section~\s{3.2}.
    }
    \label{fig:approach}
\end{figure}

\mypara{Direction Alignment Loss.}
Instead of directly maximizing the similarity between the prompts and the images, we draw inspiration from~\cite{patashnik2021styleclip,gal2022stylegan} to induce the direction between the prompt $\bm{P}^m_i$ and $\bm{P}^m_j$ to be aligned with the direction between the averaged embeddings of the reference images corresponding to \emph{a pair of categories} $a_i^m$ and $a_j^m$ in $\mathcal{A}_m$. This alignment of pairwise categories direction serves as a proxy task for guiding the prompts to learn the visual difference among images from category $a_i^m$ and $a_j^m$ (Figure~\ref{fig:approach}).

Specifically, we define the direction alignment loss $\mathcal{L}_\text{dir}$ to maximize the cosine similarity between the image direction and the prompt direction as follows:
\begin{align} \label{equ:4}
    \mathcal{L}_\text{dir}^{m}(\bm{S}^{m}_i, \bm{S}^{m}_j) = 1 - \bigl\langle \Delta_{\bm{I}}^{m} (i, j), \Delta_{\bm{P}}^{m} (i, j) \bigl\rangle.
\end{align}
Here, the image direction $\Delta_{\bm{I}}$ is defined as the difference of the averaged image embeddings between two categories of the attribute $\mathcal{A}_m$. Let $\mathfrak{X}_{k}^{m} = \frac{1}{|\mathcal{B}_k|}\sum_{y_n^m = a_k^m} E_\text{img} (\vx_n^m)$ be the averaged image embedding for category $a_k^m$; $|\mathcal{B}_k|$ is the number of images from category $a_k^m$ in each mini-batch. We denote the image direction as follows:
\begin{align}
    \Delta_{\bm{I}}^{m} (i, j) = \mathfrak{X}_{i}^{m} - \mathfrak{X}_{j}^{m}.
\end{align}
Similarly, the prompt direction $\Delta_{\bm{P}}$ is defined as the difference of the averaged prompt embeddings between two categories. Let $\mathfrak{P}_{k}^{m} = \frac{1}{|\mathcal{P}_k^m|} \sum_{\bm{P} \in \mathcal{P}^m_k} E_\text{text}(\bm{P}) $ be the averaged prompt embedding for attribute $a_k^m$. Specifically, $\mathcal{P}^{m}_{k} = \{ \bm{P} \in \mathcal{P}_\text{total} ~|~ o_m = k \}$ is a collection of prompts containing all the category combinations for other attributes given the category $a_k^m$ for attribute $\mathcal{A}_m$ (cf.~Equation~\ref{eq:set}). Finally, we denote the prompt direction as follows:
\begin{align}
    \Delta^{m}_{\bm{P}} (i, j) = \mathfrak{P}_{i}^{m} -  \mathfrak{P}_{j}^{m}.
\end{align}
By inducing the direction alignment, we aim to facilitate the prompt learning of more meaningful and nuanced differences between images from different categories.

\mypara{Semantic Consistency Loss.}
We observe that direction alignment loss alone may result in language drift~\cite{lu2020countering,lee2019countering,ruiz2022dreambooth} --- the prompts slowly lose syntactic and semantic properties of language as they only focus on solving the alignment task. To resolve this issue, we design a semantic consistency objective to regularize the training by maximizing the cosine similarity between the learning prompts and the original input prompt (see Figure~\ref{fig:approach}):
\begin{align} \label{equ:7}
    \mathcal{L}_\text{sem}^{m}(\bm{S}^{m}_i, \bm{S}^{m}_j) = \text{max} \Bigl(0, \lambda - \bigl \langle E_\text{text}(\bm{P}), E_\text{text}(\bm{T}) \bigl\rangle \Bigl)
\end{align}
where $\bm{P} \in \mathcal{P}^m_i \cup \mathcal{P}^m_j$ and $\lambda$ is a hyperparameter (see an analysis in 
Section~\s{4.3}). This loss is crucial for generating high-quality images that remain faithful to the input prompt.

\mypara{Optimization.}
Building upon $\mathcal{L}_\text{dir}^m$ and $\mathcal{L}_\text{sem}^m$, our total training loss for learning the inclusive tokens of a pair of categories in attribute $\mathcal{A}_m$ is written as follows:
\begin{align}
    \mathcal{L}_\text{pair}^{m} (\bm{S}_i^m, \bm{S}_j^m) = \mathcal{L}^{m}_\text{dir} (\bm{S}_i^m, \bm{S}_j^m) + \mathcal{L}^{m}_\text{sem} (\bm{S}_i^m, \bm{S}_j^m). 
\end{align}
At each iteration, we update the embeddings of inclusive tokens of all the categories from \emph{only one attribute} but freeze the parameters of inclusive tokens for all other attributes. The final objective during the whole learning process is: \vspace{-4mm}
\begin{align} 
    \mathcal{L}_\text{total} = {\color{blue}{\sum_{m=1}^{M}}} {\color{my_pink}{\sum_{1 \leq i < j \leq K_m}}}  \mathcal{L}_\text{pair}^{m} (\bm{S}_i^m, \bm{S}_j^m),
\end{align}
where the {\color{my_pink}{inner summation}} enumerates all pairwise categories for one attribute $\mathcal{A}_m$ at each iteration, while the {\color{blue}
{outer summation}} alters the attribute across the iteration.

\vspace{3mm}
\tocless
\subsection{Key Properties of \ourmethodbold} \label{ss_others}
\mypara{Generalizability.}
Unlike personalization methods that train the embeddings for a specific model (because they use diffusion losses~\cite{gal2022image,kumari2022multi,ruiz2022dreambooth}), \emph{the tokens learned by \ourmethod are transferable between different models.} 
We highlight two use cases for these tokens. (1) \emph{In-domain generation.} We use the user-specified prompt $\bm{T}$ to learn the inclusive tokens and then apply them back to $\bm{T}$ to generate inclusive images. (2) \emph{Train-once-for-all.} As shown in Equation~\ref{eq:set}, the newly introduced inclusive tokens do not change the original prompt $\bm{T}$, which implies that the learned tokens can be compatible with a different human-written prompt. For human face images, an example $\bm{T}$ for training can be any neutral prompt, \eg, \emph{``a headshot of a person''}. After training, inclusive tokens can be used to handle out-of-domain prompts (\eg, \emph{``a photo of a doctor''}) or facilitate different models~\cite{zhang2023adding,brooks2023instructpix2pix} in a plug-and-play manner, justifying the generalizability of our approach.

\mypara{Data, Memory, and Computational Efficiency.}
\ourmethod uses averaged image features to guide prompt learning, indicating that (1) only a few dozen images per category are sufficient, and (2) a balanced distribution across categories within an attribute is \emph{not} required. \ourmethod keeps the text-to-image model intact and only updates the inclusive tokens, allowing it to circumvent the costly back-propagation step in the diffusion model. Training with a single attribute takes approximately 5 minutes (1 A4500 GPU). In practice, we set the length\footnote{The token length used here is generalizable across the attributes we studied in this paper. See Appendix~\ref{app_ss_length} for a detailed ablation study.} ($q$ in Equation~\ref{eq:set}) of inclusive tokens to $3$
(which is less than 10KB) for all attribute categories of interest in our study. Hence, when scaling up to scenarios with multiple attributes, \ourmethod always has low memory requirements for both training and storing inclusive tokens.

\mypara{Comparison to Image Editing Methods.}
Our direction alignment loss may be reminiscent of the directional CLIP loss employed in image editing methods~\cite{gal2022stylegan,kim2022diffusionclip}. However, they are fundamentally different. First, our \ourmethod is designed to promote the inclusiveness, while image editing methods focus on single image manipulation. Second, image editing methods modify the source image according to the change in texts (from source to target), whereas \ourmethod learns prompts by leveraging changes in images from one category to another. This key difference suggests a significant distinction: the two methods are learning the task from completely different directions.
\vspace{3mm}
\tocless
\section{Experiments}
\label{s_exp}

We validate \ourmethod for inclusive text-to-image generation on various attributes and scenarios. We begin by introducing the experimental setup in
Section~\s{4.1},
then present the main results in
Section~\s{4.2},
and finally, show detailed ablation studies and applications in
Section~\s{4.3}.
Please see Appendix for additional details, results, and analyses.

\vspace{3mm}
\tocless
\subsection{Setup}
\label{exp_setup}

\begin{table*}
\begin{center}
\small
\tabcolsep 1.4pt
\centering
\caption{\small \textbf{Comparison with baseline methods with (a) single attribute and (b) multiple attributes.} Reference images are from CelebA. We use CLIP~\cite{radford2021learning} as the attribute classifier~\cite{cho2022dall,chuang2023debiasing}. \ourmethod achieves competitive results for both settings. \textbf{SD}: vanilla stable diffusion. \textbf{EI}: ethical intervention. \textbf{HPS}: hard prompt searching. \textbf{PD}: prompt debiasing. \textbf{CD}: custom diffusion. See Appendix~\ref{app_s_additional_results} for full results.}
\label{tab:celeba}
\vspace{1mm}
\begin{tabular}{l|cccccc|ccc}
\toprule
\multirow{3}{*}{\textbf{Method}} & \multicolumn{6}{c|}{\textbf{(a) Single Attribute}} & \multicolumn{3}{c}{\textbf{(b) Multiple Attributes}}\\
\cmidrule{2-10}
 & $\mathbb{D}_\text{KL}^\text{male} \downarrow $ & $\mathbb{D}_\text{KL}^\text{young} \downarrow $ & $\mathbb{D}_\text{KL}^\text{pale skin} \downarrow $ & $\mathbb{D}_\text{KL}^\text{eyeglass} \downarrow $ & $\mathbb{D}_\text{KL}^\text{mustache} \downarrow $ & $\mathbb{D}_\text{KL}^\text{smile} \downarrow $ & $\mathbb{D}_\text{KL}^{\text{male}\times\text{young}} \downarrow $ & $\mathbb{D}_\text{KL}^{\text{male} \times  \text{young} \times \text{eyeglass}} \downarrow $ & $\mathbb{D}_\text{KL}^{\text{male} \times \text{young} \times \text{eyeglass} \times \text{smile}} \downarrow $\\
\cmidrule{1-10}
SD~\cite{rombach2022high} & 0.343 & 0.578 & 0.308 & 0.375 & 0.111 & 0.134 & 0.882 & 1.187 & 1.406 \\
EI~\cite{bansal2022well} & 0.143 & 0.423 & 0.644 & 0.531 & 0.693 & 0.189 & 0.361 & 1.054 & 1.311 \\
HPS~\cite{ding2021cogview} & 1 $\times \text{10}^{-\text{5}}$ & 0.027 & 2.8 $\times \text{10}^{-\text{3}}$ & 0.371 & 0.241 & 4.4 $\times \text{10}^{-\text{3}}$  & 3.5 $\times \text{10}^{-\text{3}}$ & 0.399 & 0.476 \\
PD~\cite{chuang2023debiasing} & 0.322 & 0.131 & 0.165 & 0.272 & 0.063 & 0.146 & -- & -- & --\\
CD~\cite{kumari2022multi} & 0.309  & 0.284 & 0.074 & 0.301 & 0.246 & 0.469 & -- & -- & -- \\
\cmidrule{1-10}
\ourmethodbold & \textbf{2 $\times \text{10}^{-\text{6}}$} & \textbf{2 $\times \text{10}^{-\text{4}}$} & \textbf{0} & \textbf{2 $\times \text{10}^{-\text{4}}$} & \textbf{4.5 $\times \text{10}^{-\text{4}}$} & \textbf{2.5 $\times \text{10}^{-\text{3}}$} & \textbf{1.3 $\times \text{10}^{-\text{4}}$} & \textbf{0.061} & \textbf{0.094} \\
\bottomrule
\end{tabular}
\end{center}
\vspace{-5mm}
\end{table*}

\begin{figure*}[t]
    \centerline{\includegraphics[width=1\linewidth]{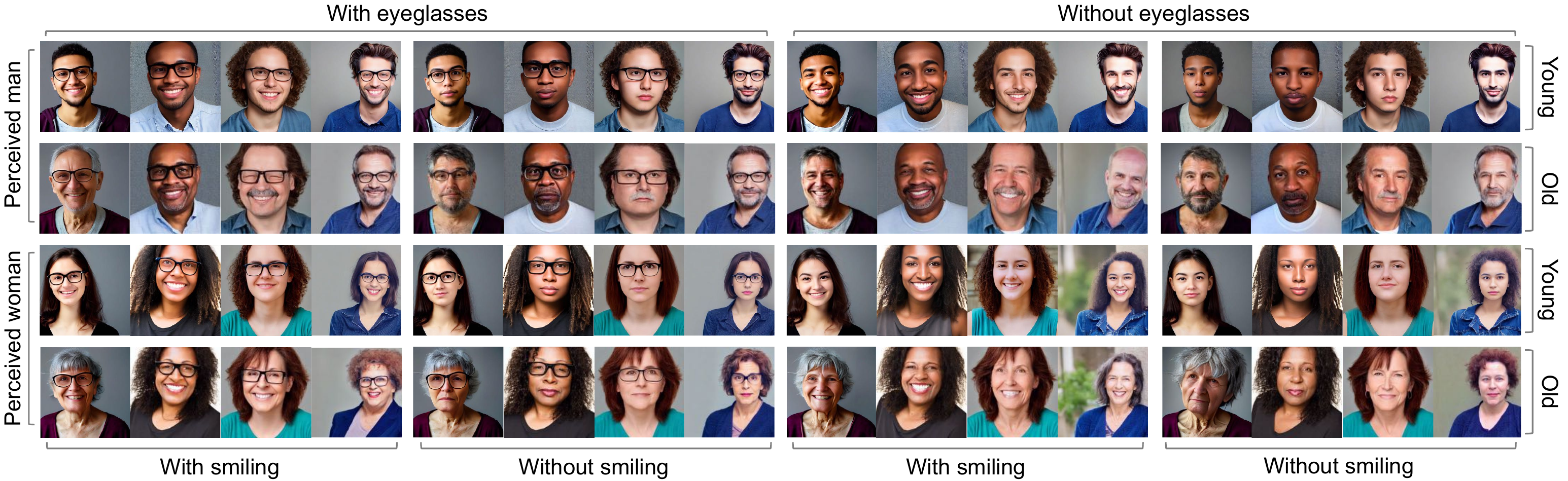}}
    \caption{\small \textbf{Qualitative results of the combination of four binary attributes} (the last column in Table~\ref{tab:celeba}). The input prompt ($\bm{T}$) is \emph{``a headshot of a person''}. By using the learned inclusive tokens (cf. Equation~\ref{eq:set}), \ourmethod can inclusively generate images with all attribute combinations. Images across each tuple are sampled using the same random seed. More examples are included in Appendix~\ref{app_s_additional_results}.}
    \label{fig:2222} 
    \vspace{-4mm}
\end{figure*}

\begin{figure}[t]
    \centerline{\includegraphics[width=1\linewidth]{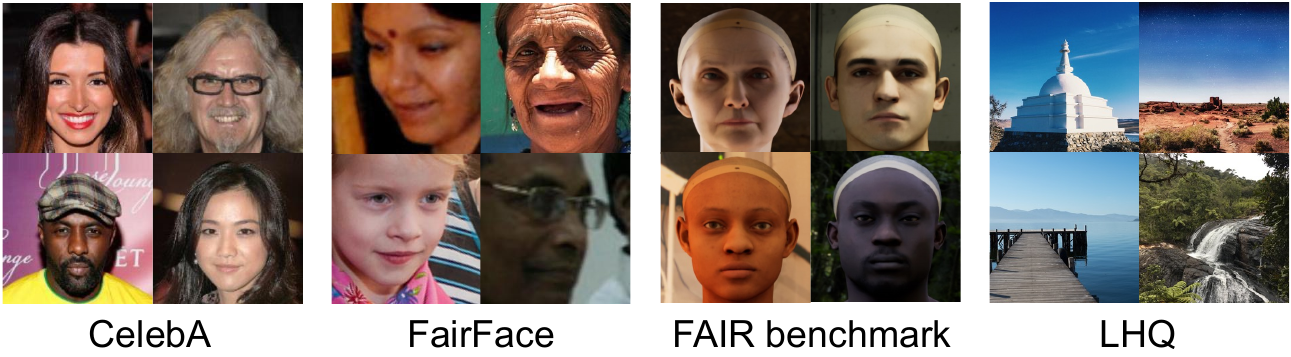}}
    \caption{\small \textbf{Examples of reference images.} CelebA~\cite{liu2015faceattributes} and FairFace~\cite{karkkainen2019fairface} are real-face datasets with different resolutions and focuses. FAIR benchmark~\cite{Feng:TRUST:ECCV2022} is a synthetic dataset used for skin tone estimation. Landscape (LHQ)~\cite{skorokhodov2021aligning} contains images from natural scenes. \ourmethod can leverage various image sources to benefit inclusive text-to-image generation for various attributes.}
    \label{fig:sample}
    \vspace{-2mm}
\end{figure}

\mypara{Datasets.}
We construct reference image sets and investigate a variety of attributes based on the following datasets. 
\textbf{(1) CelebA}~\cite{liu2015faceattributes} is a face attributes dataset and each image with $40$ binary attribute annotations. We experiment with these binary attributes and their combinations.
\textbf{(2) FAIR benchmark (FAIR)}~\cite{Feng:TRUST:ECCV2022} is a recently proposed synthetic face dataset used for skin tone estimation. Following~\cite{Feng:TRUST:ECCV2022}, we use the ground-truth albedos to classify each facial crop into one of six skin tone levels~\cite{fitzpatrick1988validity} and use FAIR for inclusiveness on skin tone type.
\textbf{(3) FairFace}~\cite{karkkainen2019fairface}\footnote{We note that, while the FairFace dataset contains race categories, we focus instead on skin tone in this study. This is because skin tone is more readily inferable from pixels, whereas racial identities are better understood as social concepts that are neither immutable nor biological in nature~\cite{browne2015dark,crawford2021atlas,ray2022critical,andrews2023ethical}; furthermore, phenotypic variation of skin tone within racial identification groups is well documented~\cite{monk2021unceasing}.} 
contains face images with annotations for $2$ perceived gender and $9$ perceived age categories.
\textbf{(4) Landscapes HQ (LHQ)}~\cite{skorokhodov2021aligning} provides unlabeled natural scene images. 
With the annotation tool from \cite{wang2022exploring}, each image can be labeled with $6$ quality (\eg, colorfulness, brightness) and $6$ abstraction (\eg, scary, aesthetic) attributes. Figure~\ref{fig:sample} shows example images.

\mypara{Experimental Protocols.}
We only require that a reference image set captures a marginal distribution for each attribute (cf. 
Section~\s{3.2}). 
Note that, while images from CelebA and FairFace are annotated with multiple attributes, we use only the attribute label for each target category but not others. We randomly select $25$ reference images per category as our default setting (and ablate it in
Section~\s{4.3}). 
For attribute settings, we consider \emph{single binary attribute}, \emph{multi-category attributes}, and \emph{multiple attributes} in the domains of human faces and scenes. We study both in-domain and train-once-for-all generations (cf.
Section~\s{3.3}) and further provide qualitative and quantitative analyses for each setup.

\mypara{Quantitative Metrics.}
We use two metrics to quantify distribution diversity and image quality.
(1) \emph{Distribution Discrepancy} ($\mathbb{D}_\text{KL}$). 
Following~\cite{cho2022dall, chuang2023debiasing}, we use the CLIP model to predict the attributes in the images. For attributes that CLIP might be erroneous, we leverage pre-trained classifiers~\cite{karkkainen2019fairface} combined with human evaluations. Specifically, for skin tone, which is extreme difficult to obtain an accurate scale~\cite{Shades, Google, howard2021reliability}, we adopt the most commonly used Fitzpatrick skin type~\cite{chardon1991skin} combined with off-the-shelf models~\cite{Feng:TRUST:ECCV2022} for evaluation.
(2) \emph{FID.} We report the FID score~\cite{heusel2017gans,parmar2022aliased} (FFHQ~\cite{karras2019style}) to measure the image quality. 
Please see Appendix~\ref{app_s_additional_ablation} for more details.

\mypara{Baselines.}
We compare \ourmethod to the following methods.
(1) \emph{Stable Diffusion} (SD)~\cite{rombach2022high} without any modification.
(2) \emph{Ethical Intervention} (EI)~\cite{bansal2022well} that edits the prompt by adding attribute-related interventions. 
(3) \emph{Hard Prompt Searching} (HPS)~\cite{ding2021cogview} that directly expresses the desired attribute category in the prompt.
(4) \emph{Prompts Debiasing} (PD)~\cite{chuang2023debiasing} that calibrates the bias in the text embedding by using the attribute category names. 
(5) \emph{Custom Diffusion} (CD)~\cite{kumari2022multi} that fine-tunes the text-to-image model with reference images based on Textual Inversion~\cite{gal2022image,ruiz2022dreambooth}.

\mypara{Implementation Details.}
We use Stable Diffusion~\cite{rombach2022high} (sd-v1-4) as the base model for all methods and show compatibility with ControlNet~\cite{zhang2023adding} and InstructPix2Pix~\cite{brooks2023instructpix2pix}. \ourmethod is model agnostic as long as they take token embeddings as the inputs. We set $\lambda=0.8$ in $\mathcal{L}_\text{sem}$ across all experiments and show that $\lambda$ can be robustly selected according to the prior knowledge (see 
Section~\s{4.3}).
All the inclusive tokens are initiated as zero vectors\footnote{We investigated other options such as random initialization but did not see notable differences in both generation quality and training speed.}. We set the length of the inclusive tokens to $3$ in all experiments.
\emph{There is no additional hyper-parameter in our framework.} The total number of the parameters for the inclusive tokens that need to be optimized is $\sum^{M}_{m=1} K_m \times 3 \times 768$, where $M$ is the number of attributes, $K_m$ is the category number for attribute $m$, and $768$ is the dimension of the embedding ($e$ in Equation~\ref{eq:set}). We train the models with $30$ epochs on a batch size of 16 and a learning rate of $0.01$. During training, we leverage image augmentations used in the CLIP image encoder.

\begin{figure}[t]
    \centerline{\includegraphics[width=1\linewidth]{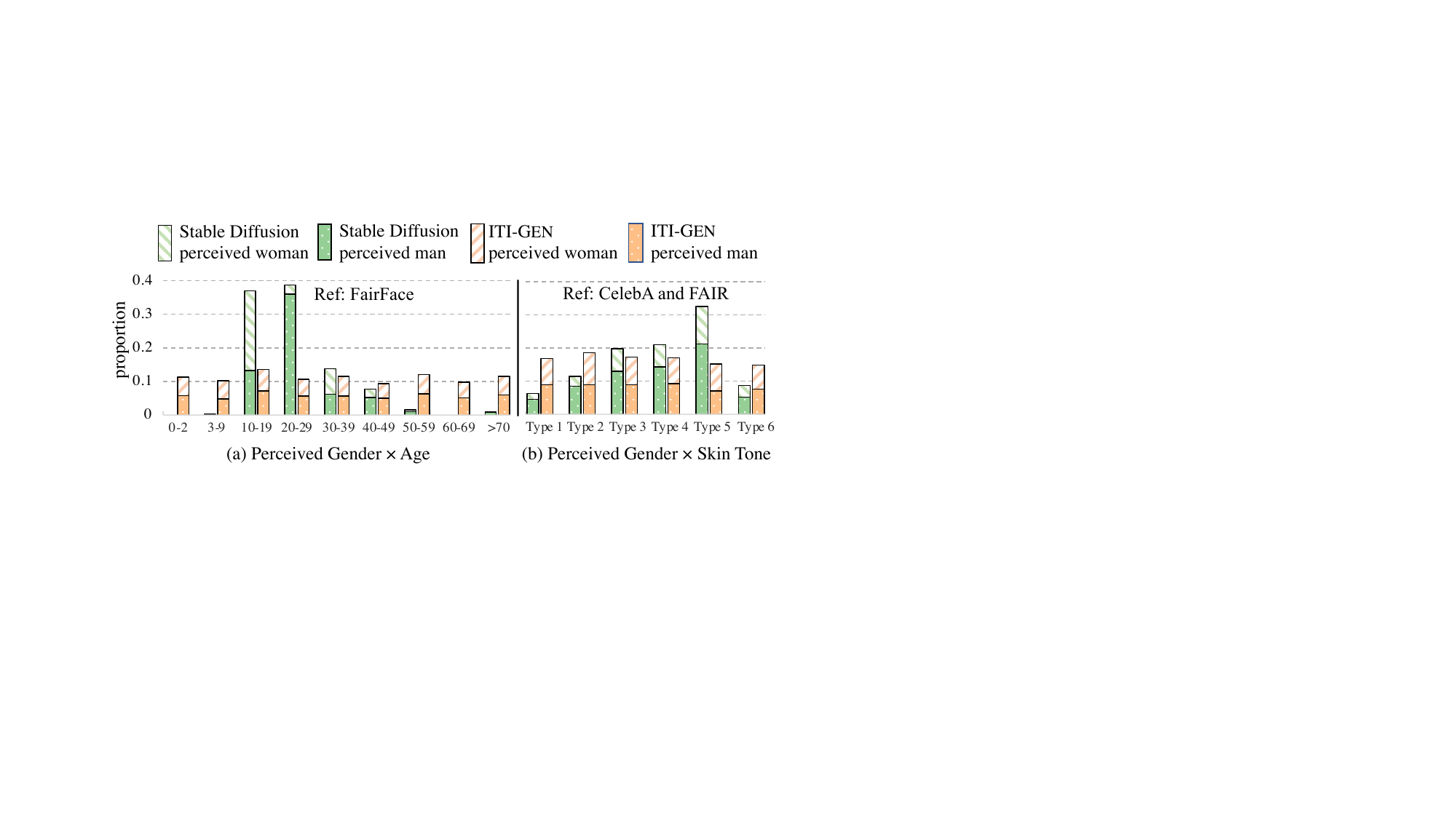}}
    \caption{\small \textbf{Multi-category distribution} with ``\emph{a headshot of a person}''. For a reliable evaluation, the results of (a) are evaluated using classifiers in~\cite{karkkainen2019fairface}, and (b) are evaluated using existing models~\cite{chardon1991skin,Feng:TRUST:ECCV2022}. The generated images from \ourmethod are more uniformly distributed across different sub-groups than the baseline Stable Diffusion. See Figure~\ref{fig:multi_category_qual} for qualitative results.}
    \label{fig:multi_category_hist}
    \vspace{-3mm}
\end{figure}

\vspace{3mm}
\tocless
\subsection{Main Results}
\label{ss_results}
\mypara{Single Binary Attribute.} To demonstrate the capability of \ourmethod to sample images with a variety of face attributes, we construct $40$ distinct reference image sets based on attributes from CelebA~\cite{liu2015faceattributes}. Each represents a specific binary attribute and contains an equal number of images ($50\%$) for the positive and negative categories\footnote{We found that different ratios do not lead to notable differences. We provide an analysis of learning with imbalanced data in Appendix~\ref{app_ss_imbalanced}.}. Table~\ref{tab:celeba}(a) shows a comparison to state-of-the-art methods. We evaluate $5$ text prompts --- ``\emph{a headshot of a \{person, professor, doctor, worker, firefighter\}}'' --- and sample $200$ images per prompt for each attribute, resulting in $40$K generated images. We highlight the averaged results across $5$ prompts of $6$ attributes. We provide complete results in Appendix~\ref{app_ss_single}. \ourmethod achieves near-perfect performance on balancing each binary attribute, justifying our motivation: using separate inclusive tokens is beneficial in generating images that are uniformly distributed across attribute categories.

\mypara{Multiple Attributes.}
Given multiple reference image sets (each captures the marginal distribution for an attribute), can \ourmethod generate diverse images across any category combination of the attributes? We provide an affirmative answer and present results in Table~\ref{tab:celeba}(b) and Figure~\ref{fig:2222}. As we observe, \ourmethod produces diverse and high-quality images with significantly lower distribution discrepancies compared to baseline methods. We attribute this to the aggregation operation of inclusive tokens (Equation~\ref{eq:set}), allowing \ourmethod to disentangle the learning of different inclusive tokens with images in marginal distributions.

\begin{figure}[t]
  \centerline{\includegraphics[width=1\linewidth]{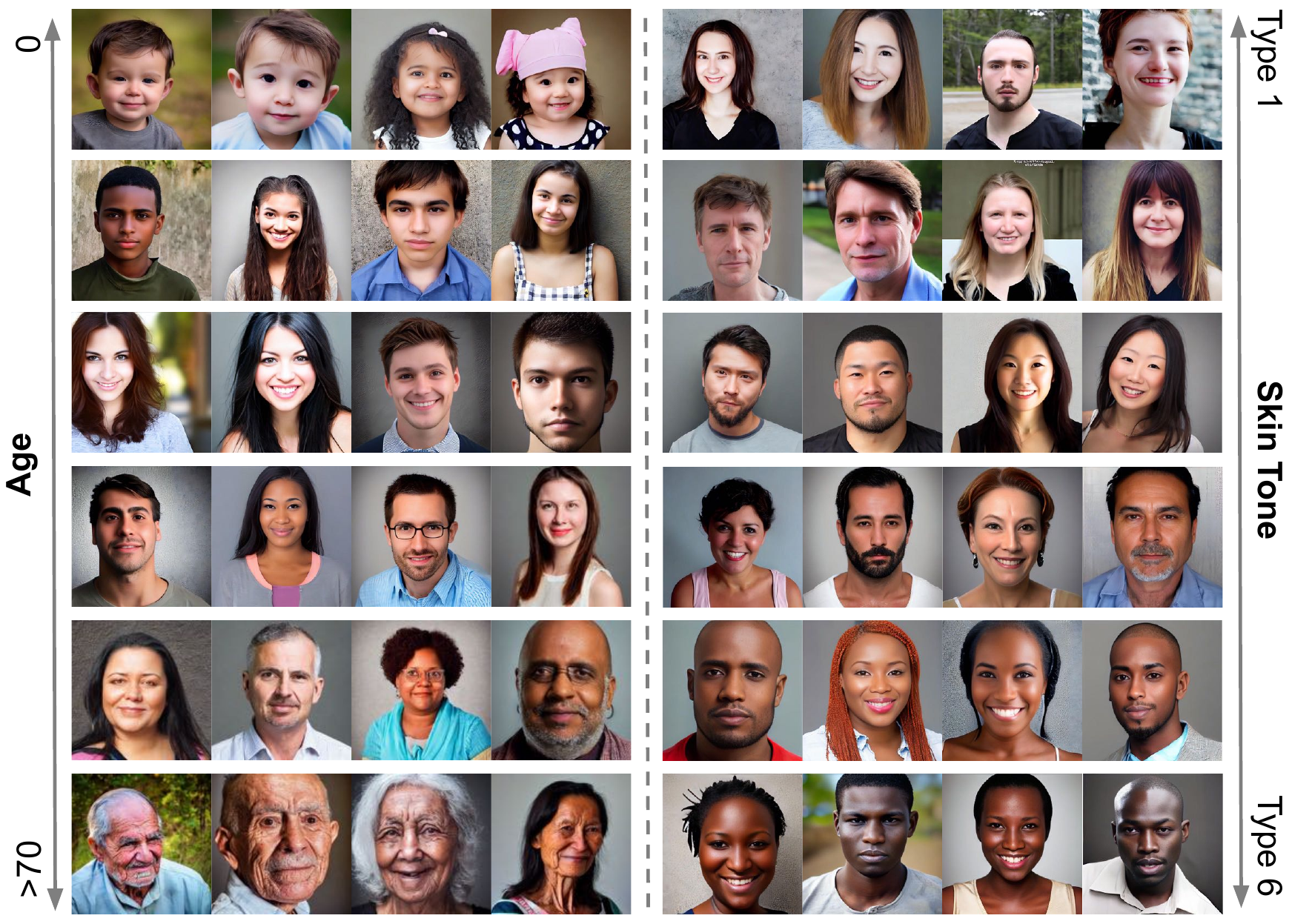}}
   \caption{\small \textbf{Results of \ourmethodbold on multi-category attributes} for Gender$\times$Age (Figure~\ref{fig:multi_category_hist}(a)) and Gender$\times$Skin Tone (Figure~\ref{fig:multi_category_hist}(b)). Examples are randomly picked with ``\emph{a headshot of a person}''.}
\label{fig:multi_category_qual}
\vspace{-3mm}
\end{figure}

\begin{figure}[t]
  \centerline{\includegraphics[width=1\linewidth]{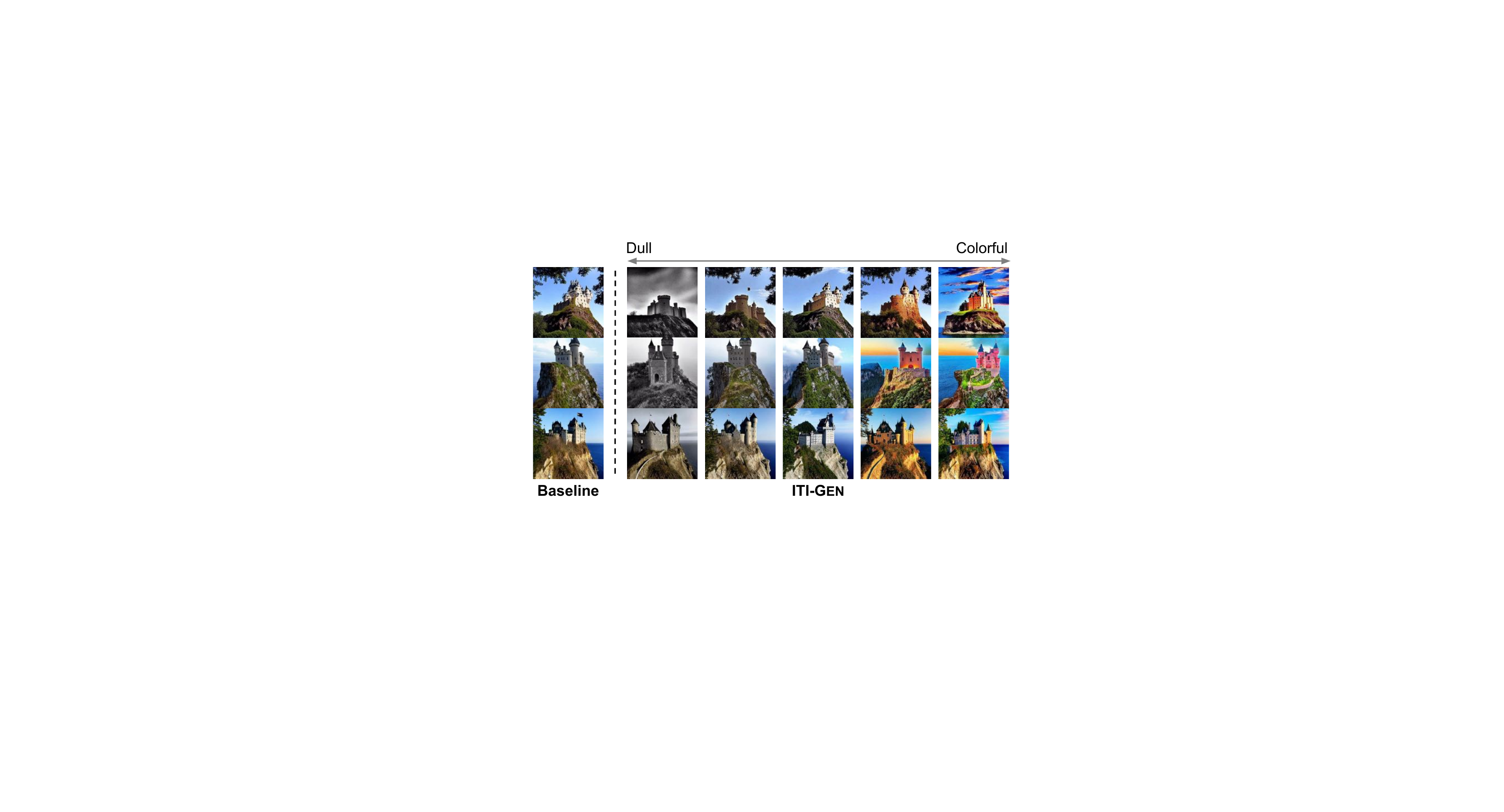}}
   \caption{\small \textbf{\ourmethodbold with perception attributes on scene images.}
   The tokens of ``colorfulness'' are trained with ``\emph{a photo of a natural scene}'' and applied to ``\emph{a castle on the cliff}'' in this example (\emph{train-once-for-all} in
   Section~\s{3.3}). \ourmethod (right) enables the baseline Stable Diffusion (left) to generate images with different levels of colorfulness. Same seed for each row. Better viewed in color. See Appendix~\ref{app_ss_other_domains} for results of other attributes, \eg, scary, brightness.}
\label{fig:scene} 
\vspace{-2mm}
\end{figure}

\mypara{Multi-Category Attributes.}
We further investigate multi-category attributes including perceived age and skin tone.
Specifically, we consider two challenging settings: (1) Perceived Gender $\times$ Age (Figure~\ref{fig:multi_category_hist}(a)), and (2) Perceived Gender $\times$ Skin Tone (Figure~\ref{fig:multi_category_hist}(b)).
\ourmethod achieves inclusiveness across all setups, especially on extremely underrepresented categories for age ($<$ 10 and $>50$ years old in Figure~\ref{fig:multi_category_hist}(a)).
More surprisingly (Figure~\ref{fig:multi_category_hist}(b)), \ourmethod can leverage synthetic images (from FAIR) and jointly learn from different data sources (CelebA for gender and FAIR for skin tone), demonstrating great potential for bootstrapping inclusive data generation with graphics engines.

\mypara{Other Domains.}
Besides human faces, we apply \ourmethod to another domain: scene images. We claim that the inclusive text-to-image generation accounts for attributes from not only humans but also scenes, objects, or even environmental factors. Specifically, we use images from LHQ~\cite{skorokhodov2021aligning} as guidance to learn inclusive tokens and generate images with diverse subjective perception attributes. As illustrated in Figure~\ref{fig:scene}, \ourmethod can enrich the generated images to multiple levels of colorfulness\footnote{Note that the subjective attributes we explore here are different from artistic styles (\eg, painting, cartoon) in image-to-image translation (\eg, \cite{gal2022stylegan}). Understanding the attributes related to \emph{quality} and \emph{look} of images may be intuitive for humans but remain non-trivial for generative models.}, justifying the generalizability of our method to the attributes in different domains.

\begin{figure}[t]
    \centerline{\includegraphics[width=1.0\linewidth]{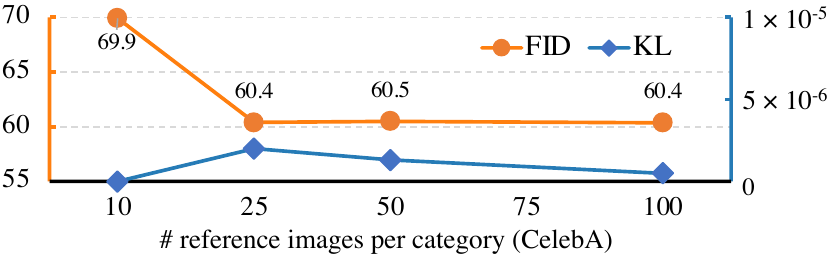}}
    \vspace{-0.5mm}
    \caption{\small \textbf{Ablation on the quantity of reference images.} More reference images ($>10$) help possibly due to more diversity and less noise. \ourmethod is robust in the low data regime (Section~\s{3.3}).}
    \label{fig:quan_refer_img}
    \vspace{-1mm}
\end{figure}

\begin{table}[t]
\small
\centering
\tabcolsep 9.5pt
\caption{\small \textbf{Ablation on reference image sources and $\mathcal{L}_\text{sem}$.} \ourmethod produces lower FID than the baseline Stable Diffusion. Semantic consistency loss $\mathcal{L}_\text{sem}$ plays a key role in quality control.}
\label{tab:fid}
\vspace{1mm}
\begin{tabular}{lccr}
\toprule
\multicolumn{1}{l}{\textbf{Method}}     & \multicolumn{1}{c}{\textbf{Source}}    & \textbf{$\mathcal{L}_\text{sem}$} & \multicolumn{1}{r}{\textbf{FID}$\downarrow$}    \\
\cmidrule{1-4}
Baseline~\cite{rombach2022high}  & \multicolumn{1}{c}{--}     &  --    &       67.40 \\
\cmidrule{1-4}
\multirow{7}{*}{\ourmethod}  & \multirow{2}{*}{CelebA~\cite{liu2015faceattributes}}   & {\color{ForestGreen}{\cmark}}       & \textbf{60.38}      \\ 
                             &                                                          & {\color{purple}{\xmark}}       & {\color{gray}{{\scriptsize (+17.40)}}} 77.78  \\ 
\cmidrule{2-4}
                             & \multirow{2}{*}{FairFace~\cite{karkkainen2019fairface}} & {\color{ForestGreen}{\cmark}}       &      \textbf{55.10} \\
                             &                                                          &  {\color{purple}{\xmark}}      &    {\color{gray}{{\scriptsize (+9.01)}}} 64.11   \\
\cmidrule{2-4}
                             & \multirow{2}{*}{FAIR~\cite{Feng:TRUST:ECCV2022}}     & {\color{ForestGreen}{\cmark}}       &       \textbf{51.83} \\
                             &                                                       &  {\color{purple}{\xmark}}      &    {\color{gray}{{\scriptsize (+10.86)}}} 62.69   \\
\bottomrule
\end{tabular} \vspace{-3mm}
\end{table}

\vspace{3mm}
\tocless
\subsection{Ablations and Applications}
\label{ss_ablation}

\mypara{Reference Images.}
Figure~\ref{fig:quan_refer_img} illustrates the impact of the \emph{quantity} of reference images per attribute category, telling that \ourmethod can produce high-quality images using very few reference data without sacrificing inclusiveness (KL).
In addition, as indicated in Table~\ref{tab:fid}, \ourmethod consistently generates realistic images regardless of reference sources (see examples in Figure~\ref{fig:2222} and Figure~\ref{fig:multi_category_qual}). More interestingly, we found that using synthetic images (\ie, FAIR~\cite{Feng:TRUST:ECCV2022}) is slightly better than real data~\cite{liu2015faceattributes,karkkainen2019fairface}. 
We hypothesize that the background noise in real images degrades the quality.

\mypara{Semantic Consistency Loss $\mathcal{L}_\text{sem}$.} 
Again in Table~\ref{tab:fid}, we compare \ourmethod with and without $\mathcal{L}_\text{sem}$.
With the help of the semantic constraint (Figure~\ref{fig:approach}), we regularize the learned embeddings not too far from the original prompt. 
We show evidence to verify this insight: the averaged CLIP similarity scores of text features between the hard prompts of $40$ attributes in CelebA and the original prompt is $0.8$ (the $\lambda$ we used), suggesting that the hyper-parameter can be robustly chosen based on prior linguistic knowledge.

\begin{figure}[t]
    \centerline{
   \includegraphics[width=1.0\linewidth]{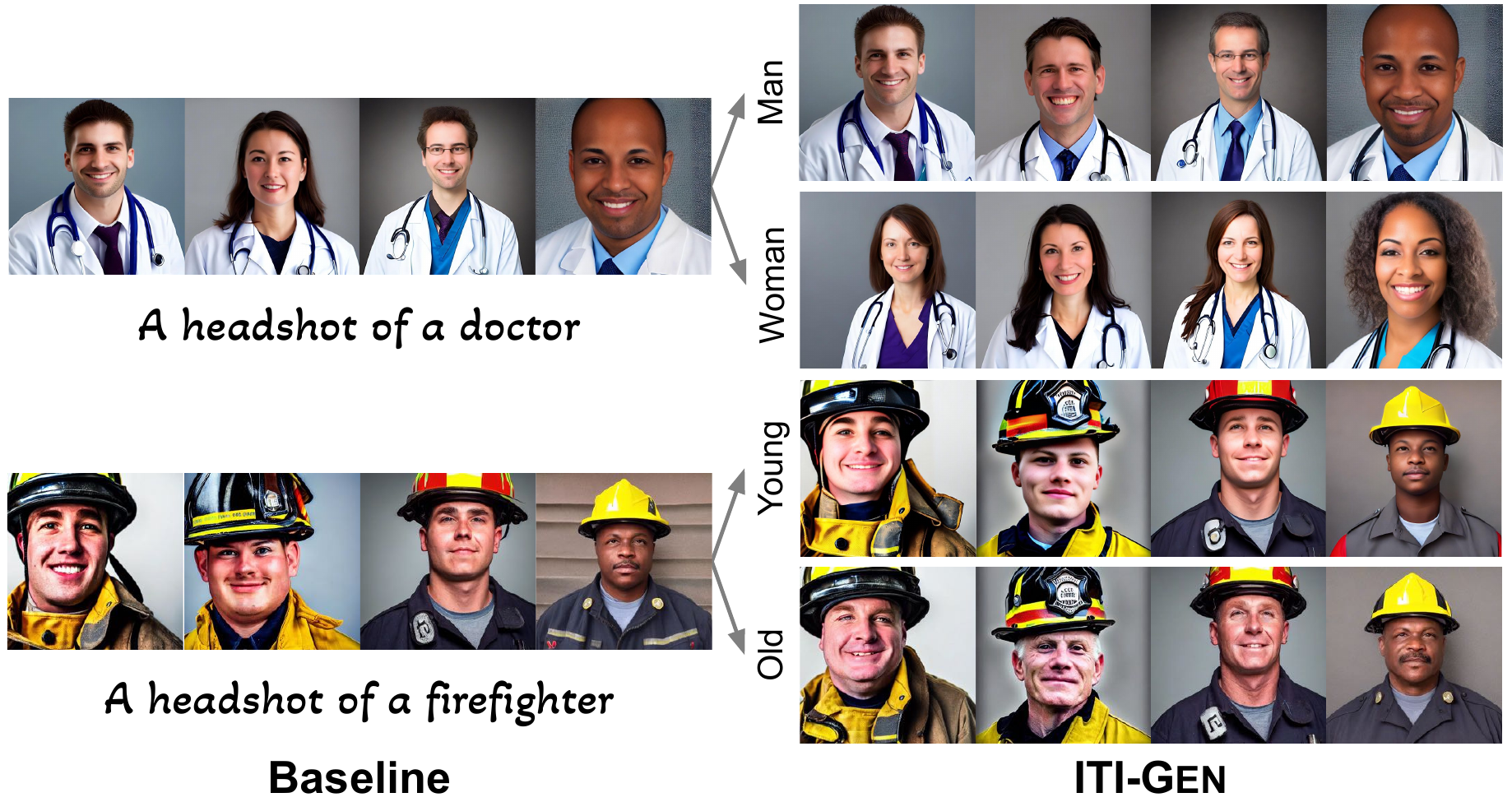}} 
   \caption{\small \textbf{Train-once-for-all generalization.} Inclusive tokens of \ourmethod trained with a neutral prompt (``\emph{a headshot of a person}'') can be applied to out-of-domain prompts in these two examples to alleviate stereotypes. See Appendix~\ref{app_ss_train_once} for more results.}
\label{fig:prepend}
\vspace{-1mm}
\end{figure}

\begin{figure}[t]
\begin{center}
   \includegraphics[width=1\linewidth]{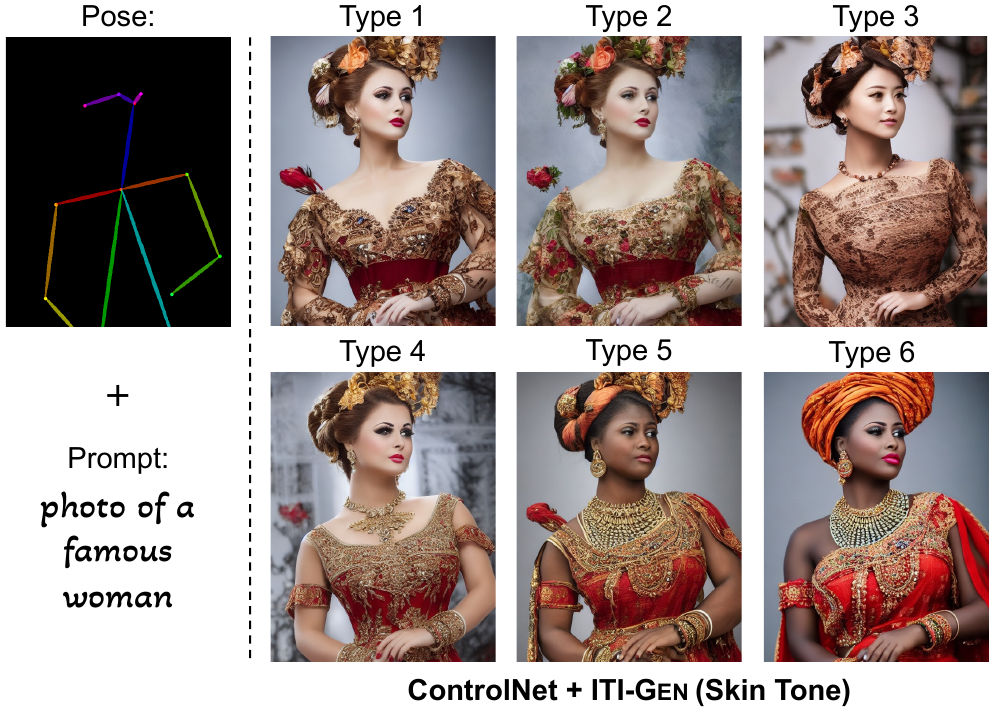}
\end{center}
\vspace{-4mm}
\caption{\small \textbf{Compatibility with models using additional conditions}, \eg, human pose (left).
\ourmethod promotes inclusiveness of ControlNet~\cite{zhang2023adding} by using the inclusive tokens of six skin tone types (right). The tokens are trained with ``\emph{a headshot of a person}'' guided by images from FAIR dataset~\cite{Feng:TRUST:ECCV2022}, and applied here in a \emph{train-once-for-all} manner 
(Section~\s{3.3}).
See Appendix~\ref{app_ss_compatible} for additional results on versatile conditions, \eg, depth, segmentation.}
\label{fig:controlNet}
\vspace{-1mm}
\end{figure}

\mypara{Train-once-for-all Generalization.}
As shown in Figure~\ref{fig:scene}, inclusive tokens can be applied to user-specified prompts in a plug-and-play manner 
(Section~\s{3.2}).
In Figure~\ref{fig:prepend}, we provide more examples of professional prompts to demonstrate the ability of train-once-for-all generation.

\mypara{Compatibility with ControlNet}~\cite{zhang2023adding}. 
\ourmethod achieves inclusiveness by learning attribute-specific prompts without modifying the original text-to-image model, potentially benefiting various downstream vision-language tasks. In Figure~\ref{fig:controlNet}, we demonstrate its compatibility with ControlNet~\cite{zhang2023adding}, a state-of-the-art model capable of conditioning on a variety of inputs beyond text. 
Interestingly, we observe an intriguing feature where the newly introduced tokens may implicitly entangle other biases or contrasts inherent in the reference image sets, such as clothing style. Nevertheless, we emphasize that disentanglement of attributes is not the primary concern of this study. \ourmethod achieves competitive results in distributional control for the \emph{intended} attributes (\eg, skin tone in Figure~\ref{fig:controlNet}) --- aggregating tokens learned from marginal distributions implicitly disentangles the \emph{known} attributes of interest.

\begin{figure}[t]
    \centerline{\includegraphics[width=1\linewidth]{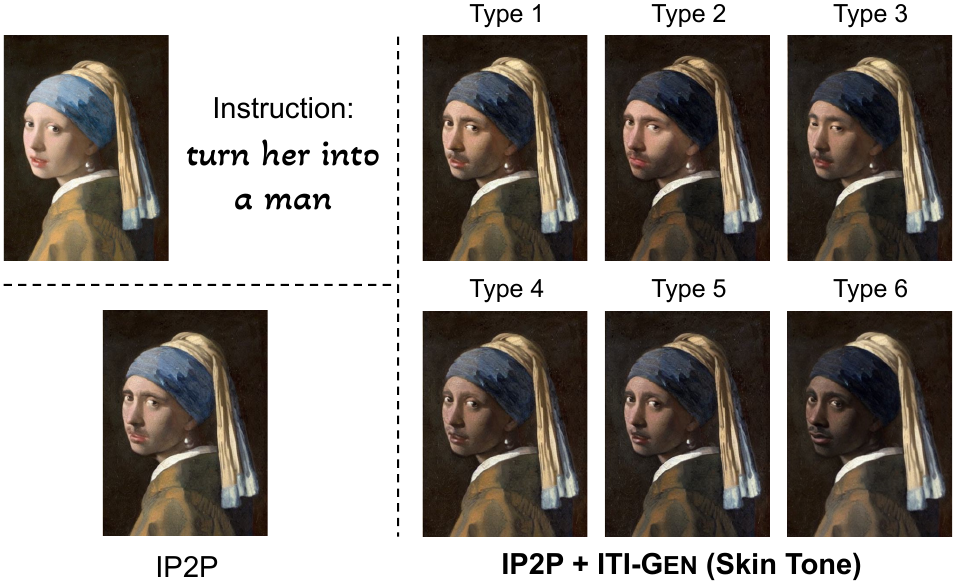}}
    \caption{\small \textbf{Compatibility with instruction-based image editing methods.} Given an image and a written instruction (top-left), InstructPix2Pix (IP2P)~\cite{brooks2023instructpix2pix} follows the instruction to edit the image (bottom-left). \ourmethod (right) enables inclusive instruction-based image editing. Similar to Figure~\ref{fig:controlNet}, the inclusive tokens used in this example are trained in a train-once-for-all manner.}
    \label{fig:ip2p}
    \vspace{-2mm}
\end{figure}

\mypara{Compatibility with InstructPix2Pix (IP2P)}~\cite{brooks2023instructpix2pix}.
Note that, achieving fully unsupervised disentanglement is a challenging task~\cite{locatello2019challenging}. Previous attempts in image generation often resort to additional supervision, either through the use of reference data~\cite{choi2020fair}, classifiers learned from a joint distribution~\cite{tan2020improving}, or even more robust controls such as instruction-based image editing~\cite{brooks2023instructpix2pix}. 
Here, we show that \ourmethod can potentially disentangle the target attribute by incorporating InstructPix2Pix~\cite{brooks2023instructpix2pix} --- to improve the inclusiveness of IP2P on the target attribute, while ensuring minimal changes to other features such as clothing and background. 
Results are shown in Figure~\ref{fig:ip2p}, telling that \ourmethod can be an effective method to condition diffusion on contrastive image sets, \eg, images taken by different cameras, art by unknown artists, and maybe even different identities of people.

\vspace{3mm}
\tocless
\section{Conclusion and Discussion}
\label{s_disc}

We present a new method for inclusive text-to-image generation. Our main contribution lies in a new direction: \emph{leveraging readily available reference images to improve the inclusiveness of text-to-image generation.} This problem is timely and challenging~\cite{bianchi2023easily,bansal2022well,chuang2023debiasing,friedrich2023fair,cho2022dall}. Our key insight is learning separate token embeddings to represent different attributes of interest via image guidance. The proposed \ourmethod method is simple, compact, generalizable, and effective on various applications. Specifically, \ourmethod has several advantages: (1) scalable to multiple attributes and different domains using relatively small numbers of images; (2) can be used in a plug-and-play manner to out-of-distribution, relatively complex prompts; (3) efficient in both training and inference; (4) compatible with the text-to-image generative models that support additional conditions or instructions. We conduct extensive experiments to verify the effectiveness of the proposed method on multiple domains, offering insights into various modeling choices and mechanisms of \ourmethod. We incorporate a broad spectrum of attributes in both human faces and scenes. We hope that our results and insights can encourage more future works on exploring inclusive data generation.

\mypara{Limitations.}
\ourmethod can handle a wide range of general attributes, such as perceived gender and skin tone, and excels in cases where ``Hard Prompt'' struggles. However, there remain several limitations. First, \ourmethod does not always provide optimal results for very subtle facial attributes (Appendix~\ref{app_ss_single}) or for the combinations of highly entangled attributes (Appendix~\ref{app_ss_multiple}). 
Second, \ourmethod still requires dozens of reference images for each category as guidance. It is possible that the reference images may introduce biases or inaccuracies. One  mitigation strategy is to integrate \ourmethod with models that offer robust controls~\cite{brooks2023instructpix2pix}, such as the one highlighted in Figure~\ref{fig:ip2p}.

\mypara{Acknowledgments.}
We thank Oliver Wang, Jianjin Xu, and Or Patashnik for their feedback on the drafts of this paper.

{\small
\bibliographystyle{ieee_fullname}
\bibliography{main}
}
\clearpage

\appendix
\begin{center}
{
    \hypersetup{linkcolor=citecolor,linktocpage=true}
    \tableofcontents
}
\end{center} 

\vspace{1mm}

\section{Ethical and Social Impacts}
\label{app_s_ethical_social}

One important consideration is the potential impact on privacy and data protection. In order to generate inclusive images, \ourmethod relies on reference images that are often sourced from publicly available datasets. However, the utilization of these images raises concerns about privacy and the potential for unintended consequences, such as the misuse of personal data. It is crucial to consider ways to mitigate these risks, such as data anonymization or obtaining explicit consent from individuals whose images are used.

While \ourmethod's directional loss avoids directly measuring the distance between the prompts and the reference images, it is possible that the reference images used to represent certain attributes may themselves contain biases or inaccuracies. To address this concern, it will be important to carefully evaluate the quality and representativeness of the reference images used in the model and to develop strategies for identifying and correcting biases when they arise.

Inclusive image generation has the potential to promote greater representation and diversity in various industries, which could in turn promote greater social equality and reduce discrimination. However, it is also possible that the technology could be misused or weaponized to promote negative or harmful stereotypes. Therefore, it will be important to consider the potential risks and benefits of \ourmethod carefully for mitigating negative outcomes.

\section{Additional Related Work and Comparisons} 
\label{app_s_additional_related}
In this section, we provide a more comprehensive comparison between \ourmethod and related methods.

\mypara{Bias Mitigation Methods in Text-to-Image Generation.}
As mentioned in Section~\s{2} of the main paper, \ourmethod uses images as guidance while existing approaches focus on debiasing the prompts. Two concurrent works, Prompt Debiasing~\cite{chuang2023debiasing} and Fair Diffusion~\cite{friedrich2023fair} require the category names of the target attributes for learning fair prompts. However, we argue that, for some attributes, attribute names might be hard to specify using language (\eg, skin tone, different levels of brightness). \ourmethod learns tokens without gradient propagation through the original text-to-image models, making it more efficient in both training and deployment.

\mypara{Personalization.}
Both \ourmethod and personalized text-to-image generation methods~\cite{kumari2022multi,gal2022image} are inspired by prompt tuning~\cite{jia2022visual,lester2021power}. However, they are fundamentally different, as introduced in Section~\s{2} of the main paper. We compare with custom diffusion~\cite{kumari2022multi} in Table~\ref{tab:celeba} of the main paper mainly to provide a justification for whether the personalization methods~\cite{kumari2022multi,ruiz2022dreambooth,gal2022image} can be used in inclusive text-to-image generation.
Specifically, we attempt to provide different numbers of reference images for Custom Diffusion~\cite{kumari2022multi} and select the best results to report.
Moreover, unlike personalization methods that use diffusion losses to train the special tokens, the tokens learned by \ourmethod are generalizable between different models.

\mypara{Disentanglement.}
It is worth mentioning that the aggregation of multiple inclusive tokens learned with separate reference datasets in marginal distributions can implicitly disentangle attribute learning. However, we emphasize that the primary goal of \ourmethod is \emph{not} to achieve feature (or attribute) disentanglement~\cite{karras2019style}. Please see Section~\append{4.3} and Figure~\ref{fig:controlNet} of the main paper for a detailed discussion.

\mypara{Image-to-image Translation and Editing.}
As mentioned in Section~\s{3.3} of the main paper, the goal of our work is to promote inclusiveness or diversity but not for image editing. In image-to-image translation or editing tasks, it is required to edit the desired attribute while keeping other features of the image intact. However, we \emph{do not} have such a requirement for \ourmethod. For example, in Figure~\ref{fig:2222}, Figure~\ref{fig:multi_category_qual}, Figure~\ref{fig:prepend}, and Figure~\ref{fig:controlNet} of the main paper, while there are subtle changes to the clothing or background in the images, \emph{\ourmethod already achieves inclusiveness for the intended attribute. We show examples with the same random seeds in these figures mainly for a better comparison.}

\section{Reference Images Preparation}
\label{app_s_reference_images}

\begin{figure}[t]
    \centerline{\includegraphics[width=1\linewidth]{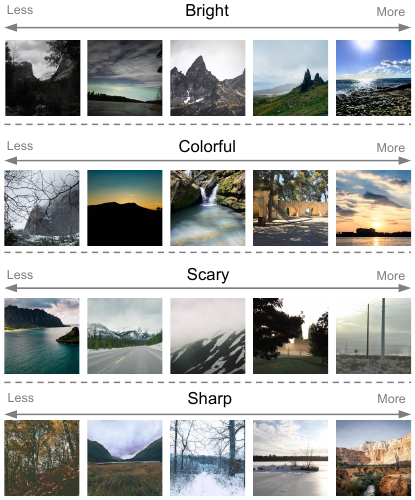}}
    \caption{\small \textbf{Examples of reference images from LHQ~\cite{skorokhodov2021aligning}.} We show randomly picked images for four attributes. Images within each category are classified into one of five groups.
    }
    \label{app_fig:lhq}
\end{figure}

In this section, we provide more details on the construction of reference image sets to complement Section~\s{4.1} of the main paper. We use the following datasets as resources.

\mypara{CelebA}~\cite{liu2015faceattributes} is a benchmarked face attributes dataset and each image with $40$ binary attribute annotations. We experiment with these binary attributes and their combinations.

\mypara{FAIR Benchmark (FAIR)}~\cite{Feng:TRUST:ECCV2022} is a recently proposed synthetic face dataset used for skin tone estimation. Specifically, we use images from the validation set containing 234 images and 702 facial crops. The validation set is released with ground-truth UV albedo maps. In order to obtain ground-truth skin tone types, we follow~\cite{Feng:TRUST:ECCV2022} to compute the Individual Typology Angle (ITA) score~\cite{chardon1991skin} of an albedo map to be the average of all pixel-wise ITA values with a pre-computed skin region area. For each image, ITA can be used to classify the skin tone type according to $6$ categories, ranging from very light (\ie, type $1$) to dark (\ie, type $6$)~\cite{chardon1991skin,del2013variations}. We randomly select $25$ images per skin tone type as the reference images.

\mypara{FairFace}~\cite{karkkainen2019fairface} contains face images with annotations for $2$ perceived gender, $9$ perceived age, and $7$ race categories. As discussed in Section~\s{4.1} of the main paper, although we value the contribution of the FairFace database to the community, we prefer using race labels and instead advocate for skin tone descriptions that recognize phenotypic diversity within broad racial categories~\cite{andrews2023ethical}. Therefore, we only use their age annotations in our experiments.

\mypara{Landscape (LHQ)}~\cite{skorokhodov2021aligning} provides unlabeled natural scene images, allowing us to extend \ourmethod to a different domain beyond human faces. With the annotation tool from \cite{wang2022exploring}, each image can be labeled with a score $s$ ranging from $0$ to $1$, with a higher value indicating a closer match to the corresponding attribute. Using this score, we classify each image into one of the five degrees of the target attribute, resulting in a multi-category attribute. Figure~\ref{app_fig:lhq} shows example reference images in the LHQ dataset. \emph{Note that, the purpose of this experiment is not to justify LHQ as a perfect resource for learning tokens for perception attributes, but to investigate the capability of our \ourmethod framework that can leverage the data from another domain as guidance.}

\section{Evaluation Metrics.}
\label{app_s_evaluation_metrics}
\mypara{Distribution Discrepancy} ($\mathbb{D}_\text{KL}$). 
Following~\cite{cho2022dall, chuang2023debiasing}, we use the CLIP model to predict the attributes in the images.
For the attributes in which every category can be accurately specified by natural language, we input the original prompt combined with different names of categories into CLIP for obtaining the attribute label.
For instance, if we want to evaluate the attribute ``male" for the images generated from ``a headshot of a person'', we construct the input text of CLIP as [``a headshot of a man", ``a headshot of a woman"].
For the attributes in which some of the categories can not be specified by natural languages, such as ``eyeglasses" and ``without eyeglasses'' (due to the issue of negative prompt), we input the text [``a headshot of a person with eyeglasses'', ``a headshot of a person''].
For attributes that CLIP might be erroneous, we leverage pre-trained classifiers~\cite{karkkainen2019fairface} combined with human evaluations. Specifically, for the skin tone, which is extremely difficult to obtain an accurate scale~\cite{Shades, Google, howard2021reliability}, we adopt the most commonly used Fitzpatrick skin type~\cite{chardon1991skin} combined with off-the-shelf models~\cite{Feng:TRUST:ECCV2022} for evaluation.

\mypara{Fréchet Inception Distance (FID)}~\cite{heusel2017gans}.
We report the FID score to measure image quality. Specifically, we use the CleanFID library~\cite{parmar2022aliased} to calculate the FID relates to statistics in FFHQ~\cite{karras2019style}.

\section{Additional Ablations and Analyses}
\label{app_s_additional_ablation}

\begin{figure}[t]
    \centerline{\includegraphics[width=1\linewidth]{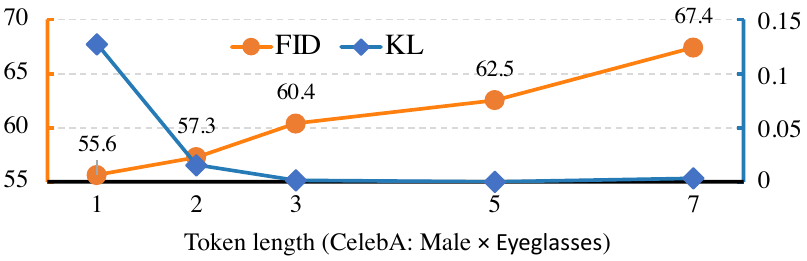}}
    \caption{\small \textbf{Ablation study on tokens length.} Using fewer tokens is not sufficient enough to capture the concepts of attributes, leading to a relatively high distribution discrepancy (\ie, KL). On the other hand, using more tokens may degrade the image quality due to language drifts (\ie, relatively high FID scores).}
    \label{app_fig:token_length}
\end{figure}

\begin{figure}[t]
    \centerline{\includegraphics[width=1.0\linewidth]{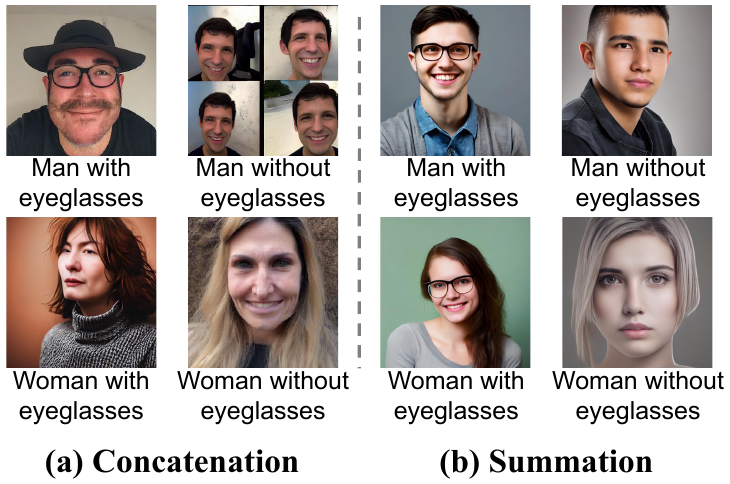}}
    \caption{\small \textbf{Concatenation \emph{vs.} summation on inclusive tokens aggregation.} We show an example of the combination of ``Male" and ``Eyeglasses" attributes. \textbf{(a)} Simply concatenating may reduce the image quality or fail to generate the images with corresponding attributes (\eg, ``Woman with eyeglasses") potentially because of the language drifts~\cite{lee2019countering,ruiz2022dreambooth}. \textbf{(b)} \ourmethod provides better results with a conceptually simpler summation.}
    \label{app_fig:ablation_aggregation}
\end{figure}

\subsection{Tokens Length}
\label{app_ss_length}
In our experiments, we set the length of inclusive tokens as $3$ ($q$ in Equation~\ref{eq:set} of the main paper). Here, we provide further analyses on the size of $q$ and show results in Figure~\ref{app_fig:token_length}. We see that fewer than $3$ tokens may hurt the performance --- cannot generate images with the desired attributes --- potentially due to less representation capacity in capturing the concepts in the reference images. On the other hand, more tokens may result in adversarial effects or collapse. We hypothesize that prepending too many tokens after the original prompts leads to language drifts~\cite{lee2019countering,ruiz2022dreambooth}. This cannot be alleviated even with the semantic consistency loss (Equation~\ref{equ:7} of the main paper) because simply forcing the two prompts with very different lengths to be close in the embedding space is ineffective.

\subsection{Tokens Aggregation}
\label{app_ss_aggregation}
As mentioned in Section~\s{3.1} of the main paper, we use summation operation to aggregate the inclusive tokens of multiple attributes to achieve permutation invariance. Here, we provide another option --- concatenation. Specifically, we ignore the positional encodings before feeding the inclusive tokens in the CLIP text encoder. Thus, the attention mechanism applied to prompt tokens is permutation invariant. Figure~\ref{app_fig:ablation_aggregation} shows comparison results. We notice that \ourmethod (with token summation) not only achieves better results than concatenation but also offers a simpler and cleaner solution for token aggregation.

\subsection{Imbalanced Reference Images}
\label{app_ss_imbalanced}
\begin{figure}[t]
    \centerline{\includegraphics[width=1.0\linewidth]{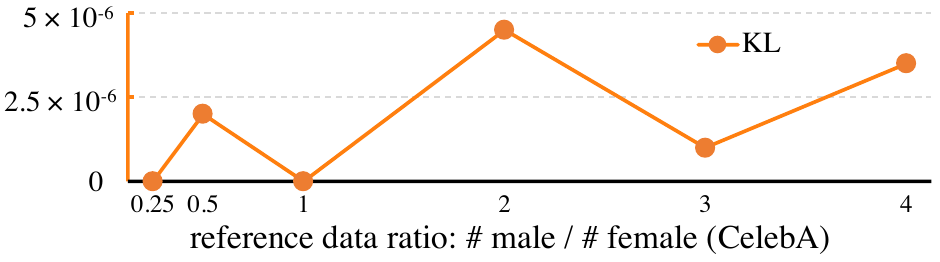}}
    \caption{\small \textbf{Ablation study on the ratio of different categories in the reference set.} We study on the perceived gender attribute in CelebA by changing the ratio of images from the ``male'' and ``female'' categories. \ourmethod is robust (\ie, with very small distribution discrepancy, KL) to the ratio of different categories in the reference image set.}
    \label{fig:ablation_imbalanced}
\end{figure}

As mentioned in Section~\s{4.1} of the main paper, we select 25 reference images per category in our experiments. We also mentioned that \ourmethod is robust to imbalanced data distributions in Section~\s{3.3}. Here, we provide additional results as evidence.
We change the ratio of ``male'' images \emph{vs.} ``female'' images for the Perceived Gender attribute in CelebA and show the results in Figure~\ref{fig:ablation_imbalanced}. \ourmethod can always generate images with nearly a balanced distribution. 

\subsection{Overlapped Reference Images}
\label{app_ss_overlapped}

\begin{figure}[t]
    \centerline{\includegraphics[width=1\linewidth]{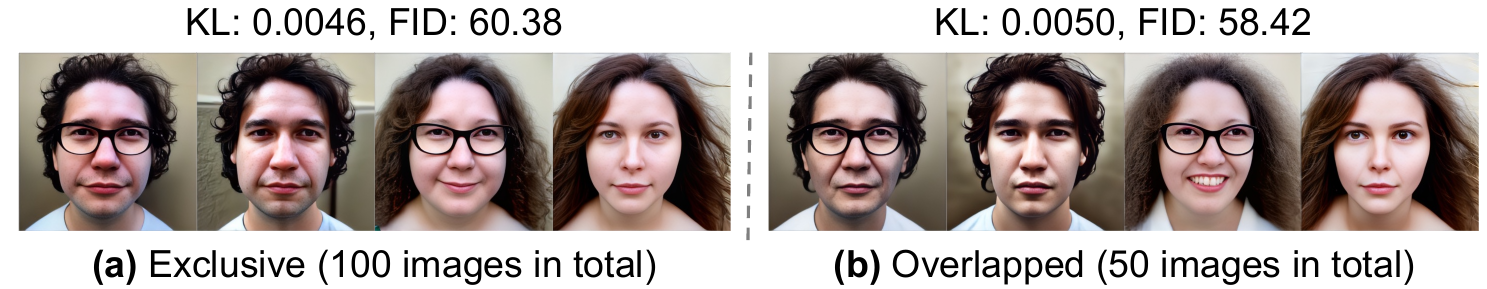}}
    \caption{\small Results of \ourmethod with (a) mutually exclusive and (b) overlapped reference images for attributes: \emph{gender}$\times$\emph{eyeglasses}.}
    \label{app_fig:overlapped}
\end{figure}

As mentioned in Section~\s{3.2} of the main paper, we need a reference dataset for each attribute.
However, this does not pose a practical issue (which seems like an all-too-exhaustive list to cover), because each reference dataset does not have to be mutually exclusive. An existing dataset (\eg, CelebA or smaller) can be divided into overlapped sub-datasets, either manually or using a classifier. To demonstrate this, we compare two settings: (a) \emph{Exclusive} --- two datasets, each containing 50 images with equal gender and eyeglasses distribution, respectively; (b) \emph{Overlapped} --- a single dataset of 50 images with equal numbers between man and woman labels, as well as with and without eyeglasses. The results in Figure~\ref{app_fig:overlapped} show that using a smaller, \emph{overlapped} dataset does not affect the performance.

\begin{figure}[t]
    \centerline{\includegraphics[width=1\linewidth]{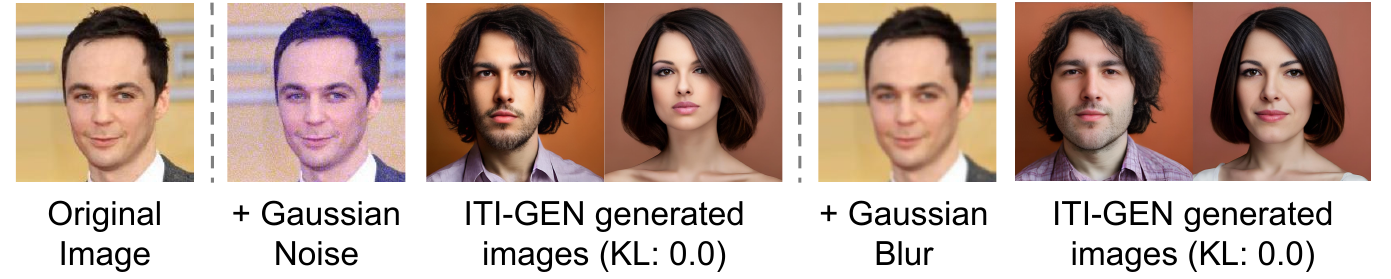}}
    \caption{\small \ourmethod with corrupted reference data. The attribute of interests is \emph{gender}.}
    \label{app_fig:noisy}
\end{figure}

\begin{figure}[t]
    \centerline{\includegraphics[width=1\linewidth]{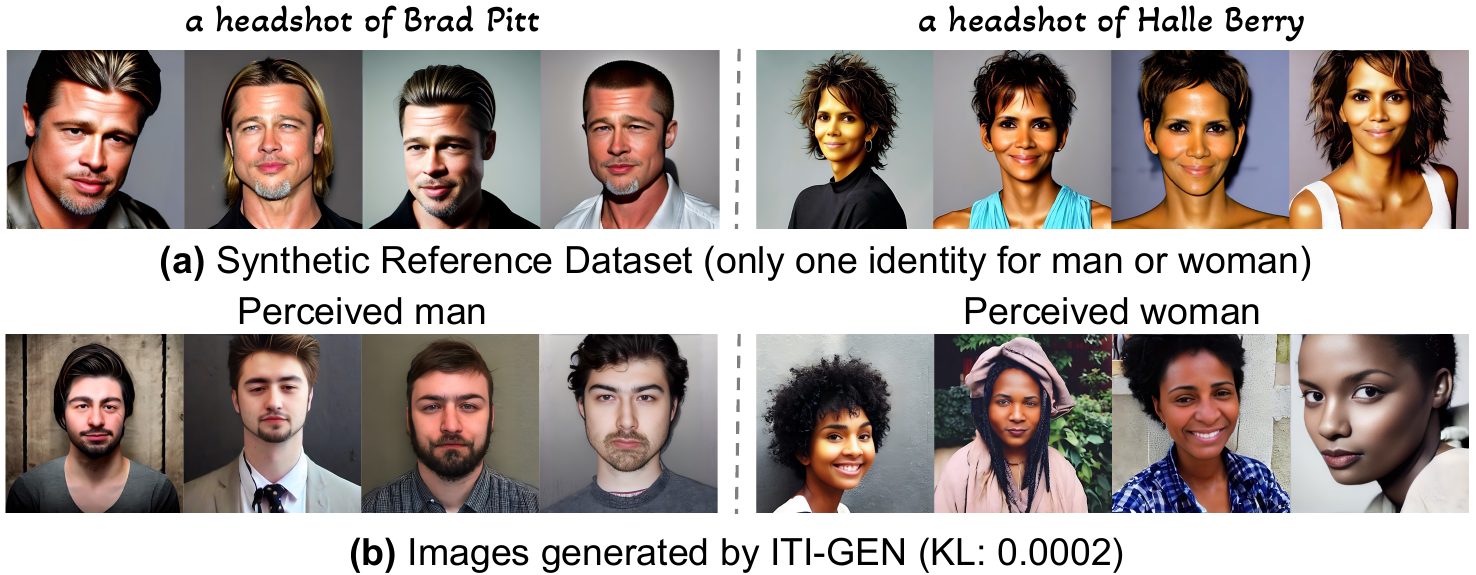}}
    \caption{\small \ourmethod can leverage less diverse reference images in (a) for inclusive generation for the \emph{gender} attribute in (b).}
    \label{app_fig:diversity}
\end{figure}

\begin{figure}[t]
    \centerline{\includegraphics[width=1\linewidth]{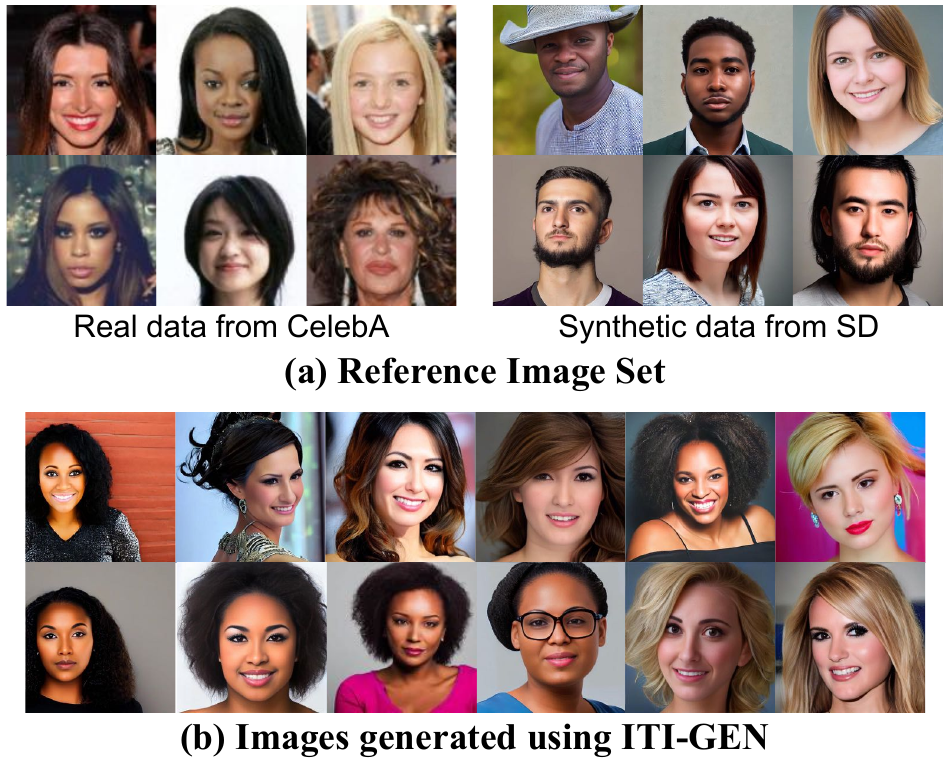}}
    \caption{\small \textbf{Qualitative results of \ourmethodbold when $K_m=1$}. When only ``female'' images are provided as the reference images (left in (a)), \ourmethod can leverage the synthetic data generated by the original prompt (``\emph{a headshot of a person}'', right in (a)), together with the real data, to construct the reference image set. By jointly using these two sources, \ourmethod learns inclusive tokens representing the concept of ``female'', which can be used to synthesize images for the desired category, as shown in (b). Section~\ref{app_ss_single_k} illustrates details.}
    \label{app_fig:ablation_one_category}
\end{figure}

\subsection{Corrupted Reference Images}
\label{app_ss_corrupted}
In this subsection, we further study whether the quality of the provided reference image strongly affects the generalization and the application of the \ourmethod. We provide the results with noisy or blurred reference images in Figure~\ref{app_fig:noisy}. We also experiment with less diverse reference images (only using the images with one identity) and show results in Figure~\ref{app_fig:diversity}. Both demonstrate the robustness of \ourmethod to the quality of reference data.

\subsection{Single Category Attribute ($K_m=1$)} 
\label{app_ss_single_k}

In the main paper, we mainly studied the attributes that have more than one category ($K_m$ is larger than $1$ in Equation~\ref{eq:set} of the main text). What if we only have the reference images from one category of the target attribute ($K_m=1$)? 
In light of our pairwise direction loss (Equation~\ref{equ:4}), there are at least two different categories needed in the reference images. Here, we show that \ourmethod can leverage the synthetic data generated by the original prompt (\eg, ``\emph{a headshot of a person}'') as an additional category to compute the directional loss for the case of $K_m = 1$.

We verify this idea by using only images of the ``female'' category from the perceived gender attribute. From Figure~\ref{app_fig:ablation_one_category}, we can observe that by leveraging the female real images and another set of synthetic images generated from ``\emph{a headshot of a person}'', \ourmethod is able to synthesize female images. Further quantitative evaluation for the images generated by \ourmethod indicates that 100\% perceived woman is obtained.

\begin{table}[t]
    \tabcolsep 4pt
    \small
    \centering
    \caption{\small \textbf{FID ($\downarrow$) comparison}. Reference images for \ourmethod are from FAIR benchmark~\cite{Feng:TRUST:ECCV2022}. \ourmethod produces lower FID than all the other baselines. \textbf{SD}: vanilla stable diffusion. \textbf{EI}: ethical intervention. \textbf{HPS}: hard prompt searching. \textbf{PD}: prompt debiasing. \textbf{CD}: custom diffusion.}
    \vspace{2mm}
    \begin{tabular}{c|c|c|c|c|c}
    \toprule
  SD~\cite{rombach2022high} & EI~\cite{bansal2022well} & HPS~\cite{ding2021cogview} & CD~\cite{kumari2022multi} & PD~\cite{chuang2023debiasing} & \ourmethodbold\\
  \midrule
      67.4 & 81.4 & 69.9 & 62.4 & 63.3 & \textbf{51.8} \\
      \bottomrule
    \end{tabular}
    \label{tab:baselines_fid}
\end{table}

\begin{figure}[t]
    \centerline{\includegraphics[width=1\linewidth]{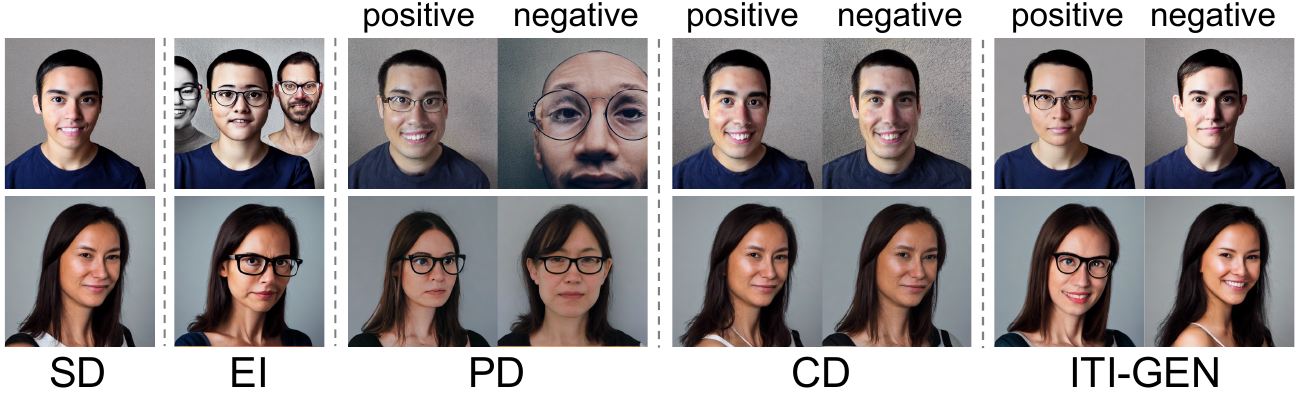}}
    \caption{\small \textbf{Visualization of different methods.} The prompt is \emph{``a headshot of a person''}. Attributes are \emph{gender} $\times$ \emph{eyeglasses}. Images across each line are sampled using the same random seed.}
    \label{fig:app_vis_baseline}
\end{figure}

\section{Additional Results}
\label{app_s_additional_results}
Due to space limitations, we only reported the results of several attributes mainly to cover the attributes relating to social factors and facial expressions in the main paper. In this section, we provide additional results and detailed comparisons to strong baseline methods.

\subsection{Qualitative and FID Results for Baselines}
\label{app_ss_qua_fid}
We only provide KL Divergence metric ($\mathbb{D}_\text{KL}$) in the main paper for different baselines. Here, we incorporate the comparisons of FID in Table~\ref{tab:baselines_fid} and visualizations in Figure~\ref{fig:app_vis_baseline} with other baselines.

\subsection{Single Binary Results}
\label{app_ss_single}
We summarize the full results of single binary attributes with CelebA~\cite{liu2015faceattributes} in Table~\ref{tab:celeba_full}. 
We compare with the baseline Stable Diffusion model~\cite{rombach2022high} and Hard Prompt Searching~\cite{ding2021cogview}, which demonstrated strong performance in many attributes (cf. Table~\ref{tab:celeba} of the main paper). 
From Table~\ref{tab:celeba_full}, we observe \ourmethod achieves the best performance in nearly all 40 attributes except some subtle facial attributes (\eg, ``Wearing Necklace'').
We use the prompt ``\emph{a headshot of a person}'' in Table~\ref{tab:celeba_full} and show qualitative results of other prompts (\eg, other occupations such as politician and musician) in Figure~\ref{app_fig:train_once_for_all}.
Furthermore, we list all the hard prompts used in our experiments in Table~\ref{app_tab:hard_specify}.

\begin{figure}[t]
    \centerline{\includegraphics[width=1\linewidth]{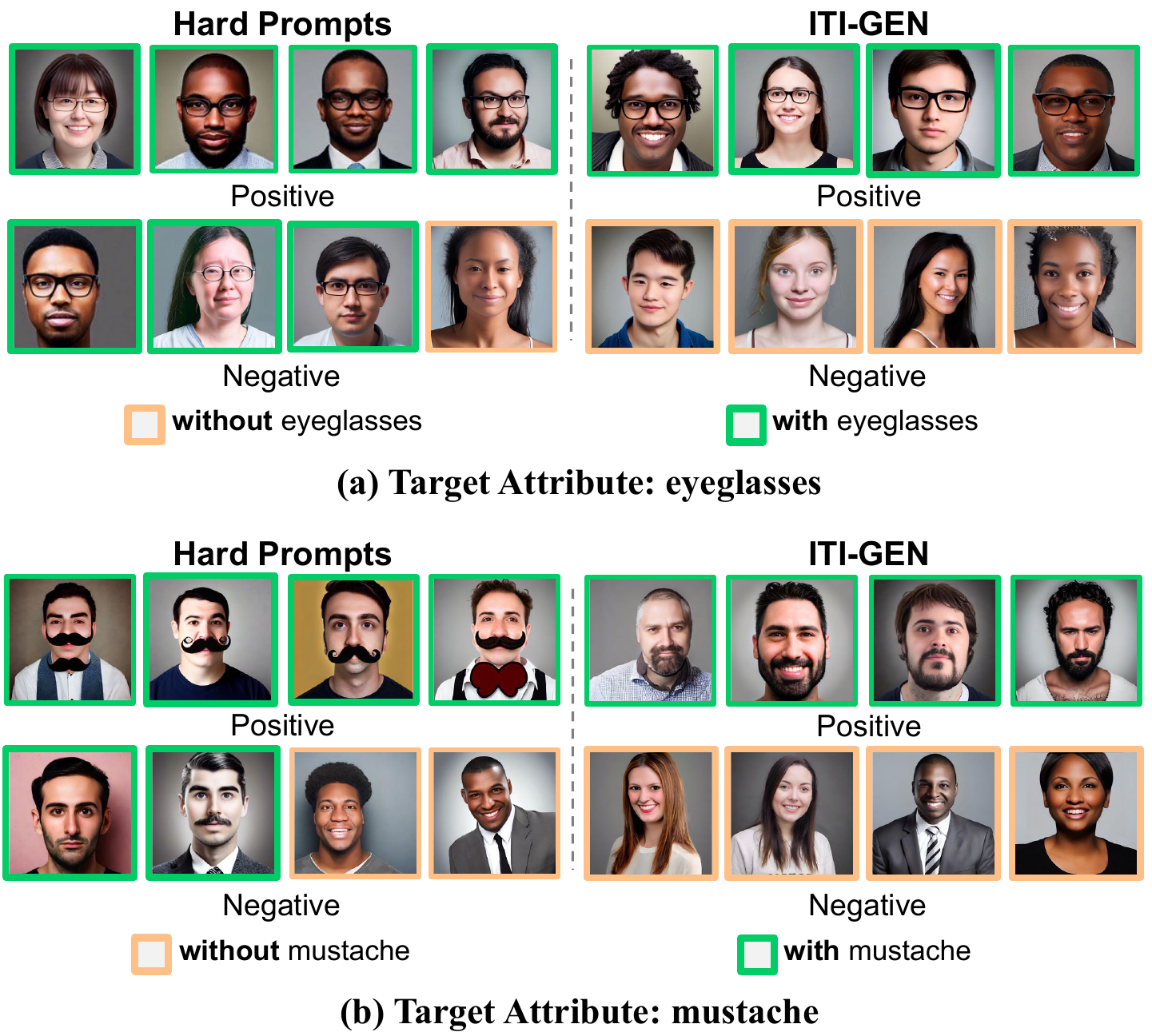}}
    \caption{\small \textbf{Challenges of (a) linguistic ambiguity and (b) model misrepresentation.} While Hard Prompts demonstrated strong capabilities in generating images with desired attributes, they cannot handle some situations. \textbf{(a)} Vanilla text-to-image models can hardly understand \emph{negative} prompts (\eg, ``not'', ``without'') possibly due to \emph{linguistic ambiguity}. \textbf{(b)} For some attributes (\eg, mustache), directly using hand prompts results in misrepresented results caused by the model bias.}
    \label{app_fig:single_binary} 
\end{figure}

\begin{figure}[t]
    \centerline{\includegraphics[width=1\linewidth]{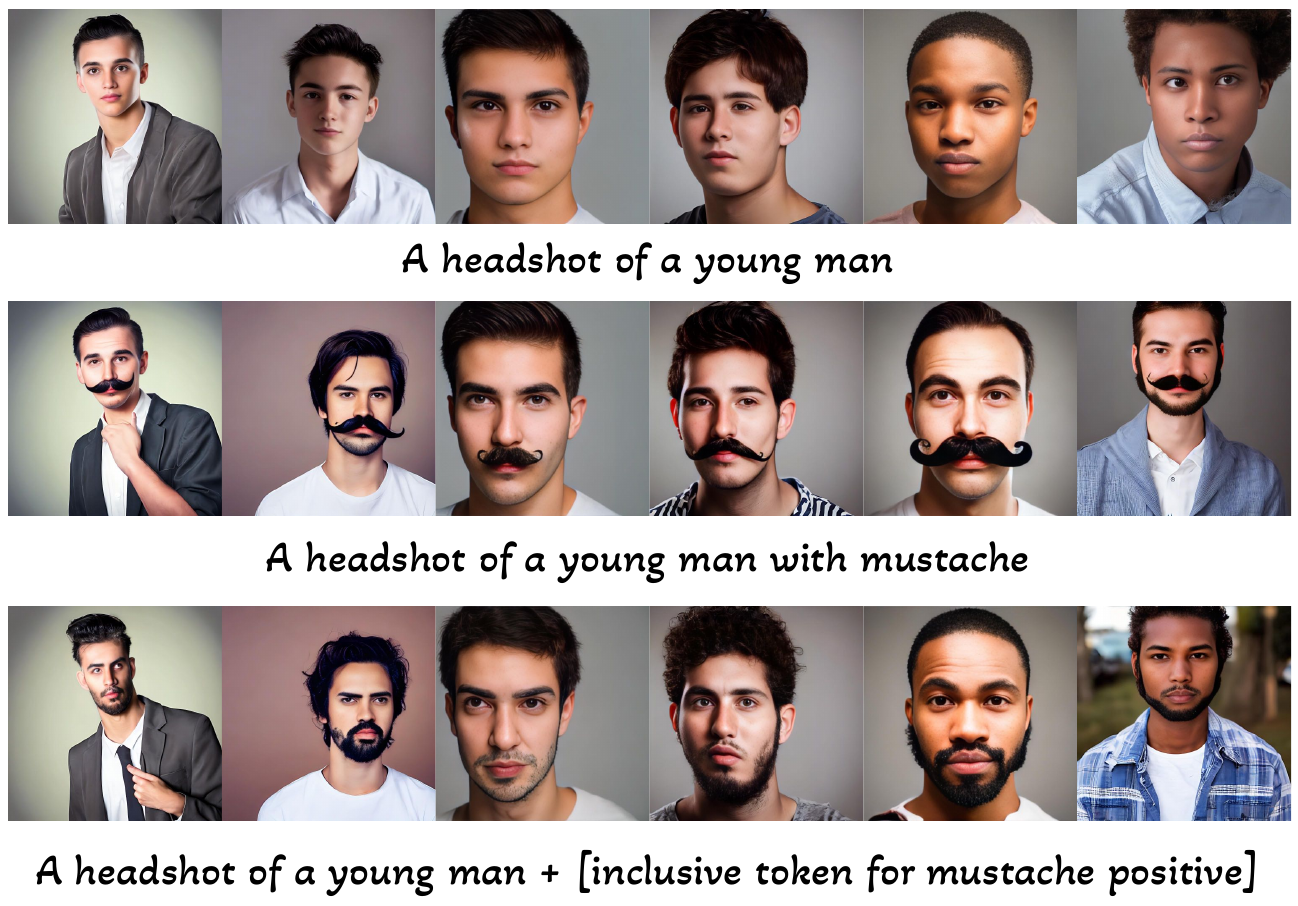}}
    \caption{\small \textbf{Compatibility of \ourmethodbold to hard prompts.} As mentioned in Section~\ref{app_ss_single} and Figure~\ref{app_fig:single_binary}, Hard Prompts show accurate results with some attributes (\eg, ``young'' and ``perceived man'' in the top row) but may result in misrepresented results for other attributes (\eg, ``mustache'' in the middle row). \ourmethod demonstrates strong compatibility with Hard Prompts to benefit a broad spectrum of attributes (bottom row).
    }
    \label{app_fig:compatible_hard} 
\end{figure}

As mentioned in Section~\s{1} of the main paper, we re-iterate that \ourmethod is designed to handle several cases (attributes) that Hard Prompts may struggle with. First, attributes with fine-grained categories may be difficult to express in language. Second, linguistic ambiguity, as shown in Figure~\ref{app_fig:single_binary} (a). Third, model misrepresentation, as illustrated in Figure~\ref{app_fig:single_binary} (b). More importantly, we argue that \ourmethod is \emph{not} to replace Hard Prompts (especially for attributes that are already can be handled by language) but to support complex prompts with multiple attributes, as illustrated in Figure~\ref{app_fig:compatible_hard}.

\begin{table}[t]
\begin{center}
\small
\tabcolsep 8pt
\caption{\small \textbf{A full comparison with baseline methods with the $40$ single attribute setting} ($\mathbb{D}_\text{KL}\downarrow $). Reference images are from CelebA~\cite{liu2015faceattributes}. Following~\cite{cho2022dall,chuang2023debiasing}, we use CLIP~\cite{radford2021learning} as the attribute classifier. \textbf{SD}: vanilla stable diffusion~\cite{rombach2022high}. \textbf{HPS}: hard prompt searching~\cite{ding2021cogview}. Given the strong capability of the existing text-to-image generative models, one can express the (most but not all) desired attributes directly using \emph{Hard Prompts}. However, it faces challenges in certain attributes and \ourmethod addresses most of these drawbacks. Please see Figure~\ref{app_fig:single_binary} for a side-by-side qualitative comparison between HPS and \ourmethod. Please see Figure~\ref{app_fig:compatible_hard} for how \ourmethod can be compatibly used with Hard Prompts.}
\label{tab:celeba_full}
\vspace{2mm}
\begin{tabular}{c|ccc}
\toprule
Attribute & SD~\cite{rombach2022high} & HPS~\cite{ding2021cogview} & \ourmethod \\
\cmidrule{1-4}
5'o Clock Shadow  & 0.02957 & \textbf{0.00847} & 0.06882 \\
Arched Eyebrows & 0.32972 & 0.04570 & \textbf{0.00892} \\
Attractive & 0.11264 & 0.07405 & \textbf{0.00000} \\
Bags Under Eyes & 0.33325 & 0.10498 & \textbf{0.01395} \\
Bald & 0.51578 & 0.22175 & \textbf{0.00892} \\
Bangs & 0.33886 & 0.19975 & \textbf{0.00000} \\
Big Lips & 0.20984 & 0.02908 & \textbf{0.00892} \\
Big Nose & 0.32423 & 0.01629 & \textbf{0.00056} \\
Black Hair & 0.35189 & 0.12539 & \textbf{0.00000} \\
Blond Hair & 0.60804 & 0.00501 & \textbf{0.00222} \\
Blurry & \textbf{0.01077} & 0.25348 & 0.09707 \\
Brown Hair & 0.41683 & 0.14207 & \textbf{0.05663} \\
Bushy Eyebrows & 0.07108 & 0.29737 & \textbf{0.02747} \\
Chubby & 0.14293 & 0.40233 & \textbf{0.00000} \\
Double Chin & 0.28637 & 0.48016 & \textbf{0.19274} \\
Eyeglasses & 0.38773 & 0.32622 & \textbf{0.00056} \\
Goatee & 0.25933 & 0.04266 & \textbf{0.00000} \\
Gray Hair & 0.65905 & 0.27921 & \textbf{0.17049} \\
Heavy Makeup & 0.39293 & 0.10989 & \textbf{0.04570} \\
High Cheekbones & 0.47875 & \textbf{0.00020} & 0.03599 \\
Male & 0.01033 & 0.00005 & \textbf{0.00000} \\
Mouth Slightly Open & 0.07030 & 0.14207 & \textbf{0.04570} \\
Mustache & 0.02013 & 0.12009 & \textbf{0.00000} \\
Narrow Eyes & 0.14968 & \textbf{0.00847} & 0.08228 \\
No Beard & 0.22442 & 0.49463 & \textbf{0.00222} \\
Oval Face & 0.39526 & 0.03158 & \textbf{0.02014} \\
Pale Skin & 0.17394 & 0.00045 & \textbf{0.00000} \\
Pointy Nose & 0.48951 & 0.02221 & \textbf{0.00000} \\
Receding Hairline & 0.31784 & 0.61526 & \textbf{0.02014} \\
Rosy Cheeks & 0.46275 & \textbf{0.03691} & 0.14987 \\
Sideburns & 0.55409 & 0.04570 & \textbf{0.02013} \\
Smiling & 0.25059 & 0.02075 & \textbf{0.00000} \\
Straight Hair & \textbf{0.08506} & 0.61526 & 0.19274 \\
Wavy Hair & 0.47663 & 0.36806 & \textbf{0.03599} \\
Wearing Earrings & 0.32029 & 0.15998 & \textbf{0.09707} \\
Wearing Hat & 0.65144 & 0.12539 & \textbf{0.01395} \\
Wearing Lipstick & 0.50658 & 0.12539 & \textbf{0.11323} \\
Wearing Necklace & 0.63600 & \textbf{0.05897} & 0.49463 \\
Wearing Necktie & 0.46687 & 0.69315 & \textbf{0.06882} \\
Young & 0.65647 & 0.00056 & \textbf{0.00000} \\
\bottomrule
\end{tabular}
\end{center}
\vspace{-8mm}
\end{table}

\begin{table*}[]
\begin{center}
\small
\tabcolsep 7pt
\fontsize{8.5}{8}\selectfont
\caption{\small \textbf{Hard Prompts used in our experiments.} Different attributes may not follow the same template and we carefully specify or express the attribute in the input prompt. The human-written hard prompts are used to generate images. Results are shown in Table~\ref{tab:celeba_full}.}
\label{app_tab:hard_specify}
\vspace{2mm}
\begin{tabular}{c|c|c}
\toprule
\textbf{Attribute} & \textbf{Positive Prompt} & \textbf{Negative Prompt} \\
\cmidrule{1-3}
5'o Clock Shadow & a headshot of a person \{with 5'o clock shadow\}. & a headshot of a person \{with no/without 5'o clock shadow\}. \\
Arched Eyebrows & a headshot of a person \{with arched eyebrows\}. & a headshot of a person \{with no/without arched eyebrows\}.\\
Attractive & a headshot of \{an attractive\} person. & a headshot of \{an unattractive\} person.\\
Bags Under Eyes & a headshot of a person \{with bags under eyes\}. & a headshot of a person \{with no/without bags under eyes\}.\\
Bald & a headshot of a person \{in bald\}. & a headshot of a person \{in no bald\}.\\
Bangs & a headshot of a person \{with bangs\}. & a headshot of a person \{with no/without bangs\}.\\
Big Lips & a headshot of a person \{with big lips\}. & a headshot of a person \{with small lips\}.\\
Big Nose & a headshot of a person \{with big nose\}. & a headshot of a person \{with small nose\}.\\
Black Hair & a headshot of a person \{with black hair\}. & a headshot of a person \{with no/without black hair\}.\\
Blond Hair & a headshot of a person \{with blond hair\}. & a headshot of a person \{with no/without blond hair\}.\\
Blurry & a headshot of a person \{in blurry\}. & a headshot of a person \{in no/without blurry\}.\\
Brown Hair & a headshot of a person \{with brown hair\}. & a headshot of a person \{with no/without brown hair\}.\\
Bushy Eyebrows & a headshot of a person \{with bushy eyebrows\}. & a headshot of a person \{with no/without bushy eyebrows\}.\\
Chubby & a headshot of a \{chubby\} person. & a headshot of a \{no chubby\} person.\\
Double Chin & a headshot of a person \{with double chin\}. & a headshot of a person \{with no/without double chin\}.\\
Eyeglasses & a headshot of a person \{with eyeglasses\}. & a headshot of a person \{with no/without eyeglasses\}.\\
Goatee & a headshot of a person \{with goatee\}. & a headshot of a person \{with no/without goatee\}.\\
Gray Hair & a headshot of a person \{with gray hair\}. & a headshot of a person \{with no/without gray hair\}.\\
Heavy Makeup & a headshot of a person \{with heavy makeup\}. & a headshot of a person \{with no/without heavy makeup\}.\\
High Cheekbones & a headshot of a person \{with high cheekbones\}. & a headshot of a person \{with low cheekbones\}.\\
Male & a headshot of a \{man\}. & a headshot of a \{woman\}.\\
Mouth Slightly Open & a headshot of a person \{with mouth slightly open\}. & a headshot of a person \{with mouth closed\}.\\
Mustache & a headshot of a person \{with mustache\}. & a headshot of a person \{with no/without mustache\}.\\
Narrow Eyes & a headshot of a person \{with narrow eyes\}. & a headshot of a person \{with no/without narrow eyes\}.\\
No Beard & a headshot of a person \{with no/without beard\}. & a headshot of a person \{with beard\}.\\
Oval Face & a headshot of a person \{with oval face\}. & a headshot of a person \{with no/without oval face\}.\\
Pale Skin & a headshot of a person \{with pale skin\}. & a headshot of a person \{with dark skin\}.\\
Pointy Nose & a headshot of a person \{with pointy nose\}. & a headshot of a person \{with no/without pointy nose\}.\\
Receding Hairline & a headshot of a person \{with receding hairline\}. & a headshot of a person \{with no/without receding hairline\}.\\
Rosy Cheeks & a headshot of a person \{with rosy cheeks\}. & a headshot of a person \{with no/without rosy cheeks\}.\\
Sideburns & a headshot of a person \{with sideburns\}. & a headshot of a person \{with no/without sideburns\}.\\
Smiling & a headshot of a person \{with smiling\}. & a headshot of a person \{with no/without smiling\}.\\
Straight Hair & a headshot of a person \{with straight hair\}. & a headshot of a person \{with no/without straight hair\}.\\
Wavy Hair & a headshot of a person \{with wavy hair\}. & a headshot of a person \{with no/without wavy hair\}.\\
Wearing Earrings & a headshot of a person \{wearing earrings\}. & a headshot of a person \{without wearing earrings\}.\\
Wearing Hat & a headshot of a person \{wearing hat\}. & a headshot of a person \{without wearing hat\}.\\
Wearing Lipstick & a headshot of a person \{wearing lipstick\}. & a headshot of a person \{without wearing lipstick\}.\\
Wearing Necklace & a headshot of a person \{wearing necklace\}. & a headshot of a person \{without wearing necklace\}.\\
Wearing Necktie & a headshot of a person \{wearing necktie\}. & a headshot of a person \{without wearing necktie\}.\\
Young & a headshot of a \{young\} person. & a headshot of \{an old\} person.\\
\bottomrule
\end{tabular}
\end{center}
\vspace{-3mm}
\end{table*}

\subsection{Multiple Attributes}
\label{app_ss_multiple}
\begin{figure*}[t]
  \centerline{\includegraphics[width=1\linewidth]{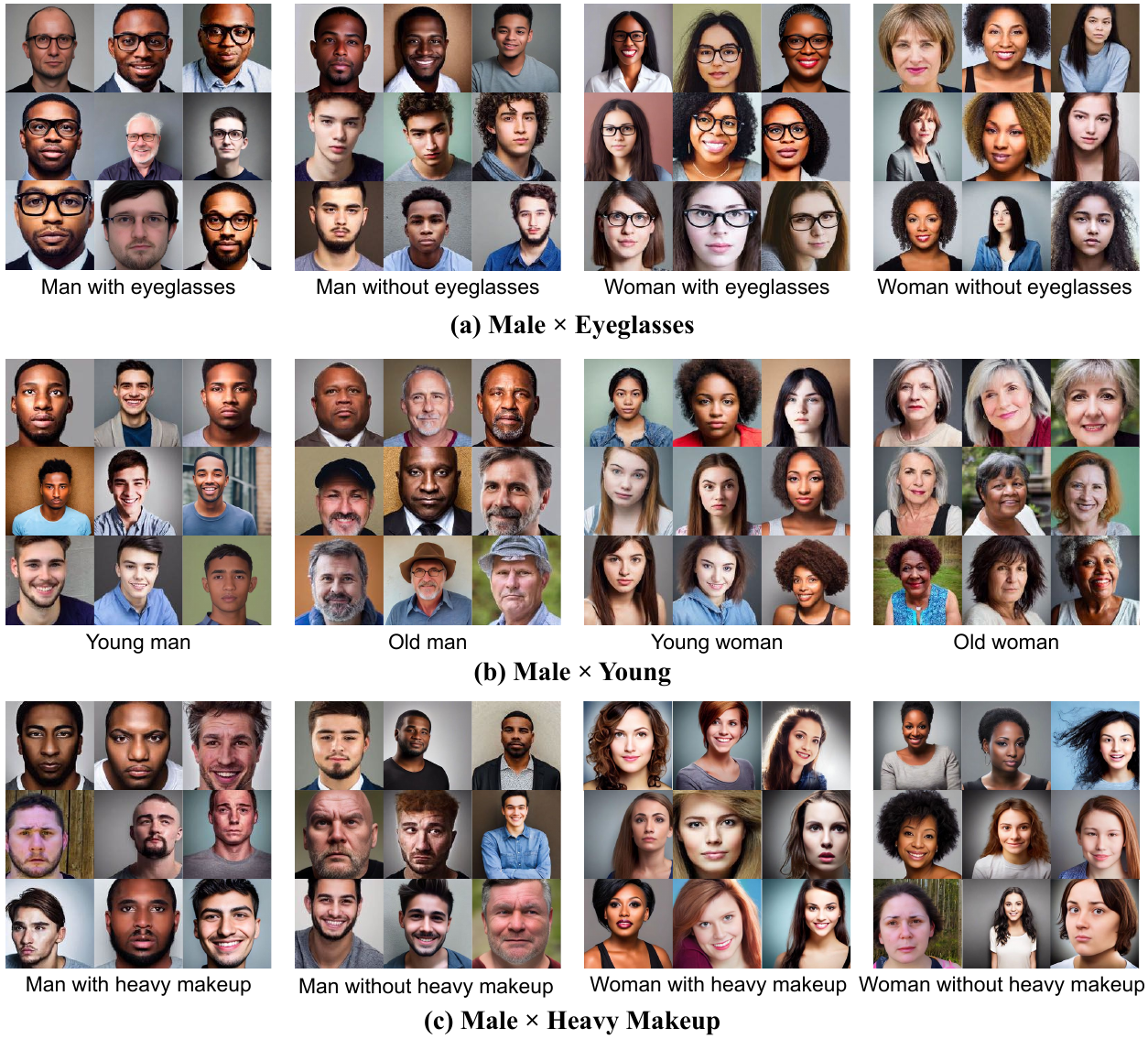}}
   \caption{\small \textbf{Additional results on multiple attributes.} We consider three settings based on the attribute co-occurrence matrix in the CelebA dataset (see Section~\ref{app_ss_multiple}). The attribute combinations in (a) and (b) are relatively less entangled between the sub-categories whereas in (c) --- a \emph{failure} case of \ourmethod --- the category ``with heavy makeup'' is heavily entangled with the category ``female'' in CelebA, which indicates that other category combinations (\eg, ``man with heavy makeup'') can rarely happen in our daily life. Therefore, the text-to-image model can hardly synthesize images with this underrepresented attribute combination.}
    \label{app_fig:multi_attributes}
\end{figure*}

We now consider multi-attribute cases and show additional results in Figure~\ref{app_fig:multi_attributes}.
To fully characterize the performance of \ourmethod, we study three additional settings based on the attribute correlation matrix from the CelebA dataset~\cite{liu2015faceattributes} (see Figure~\s{2} in \cite{torfason2017face} for the correlation matrix). Specifically, we select three attribute combinations with different levels of attribute entanglement (\ie, co-occurrence frequency) --- a higher co-occurrence value means the attribute combination is more common in daily life while a lower co-occurrence value indicates a rare case in the original CelebA dataset. Admittedly, there are several cases \ourmethod does not always generate images with a balanced distribution or faithfully generates images with specific attributes. Please see Figure~\ref{app_fig:multi_attributes} for details.

\subsection{Multi-Category Attributes}
\label{app_ss_multicategory}
\begin{figure*}[t]
  \centerline{\includegraphics[width=1\linewidth]{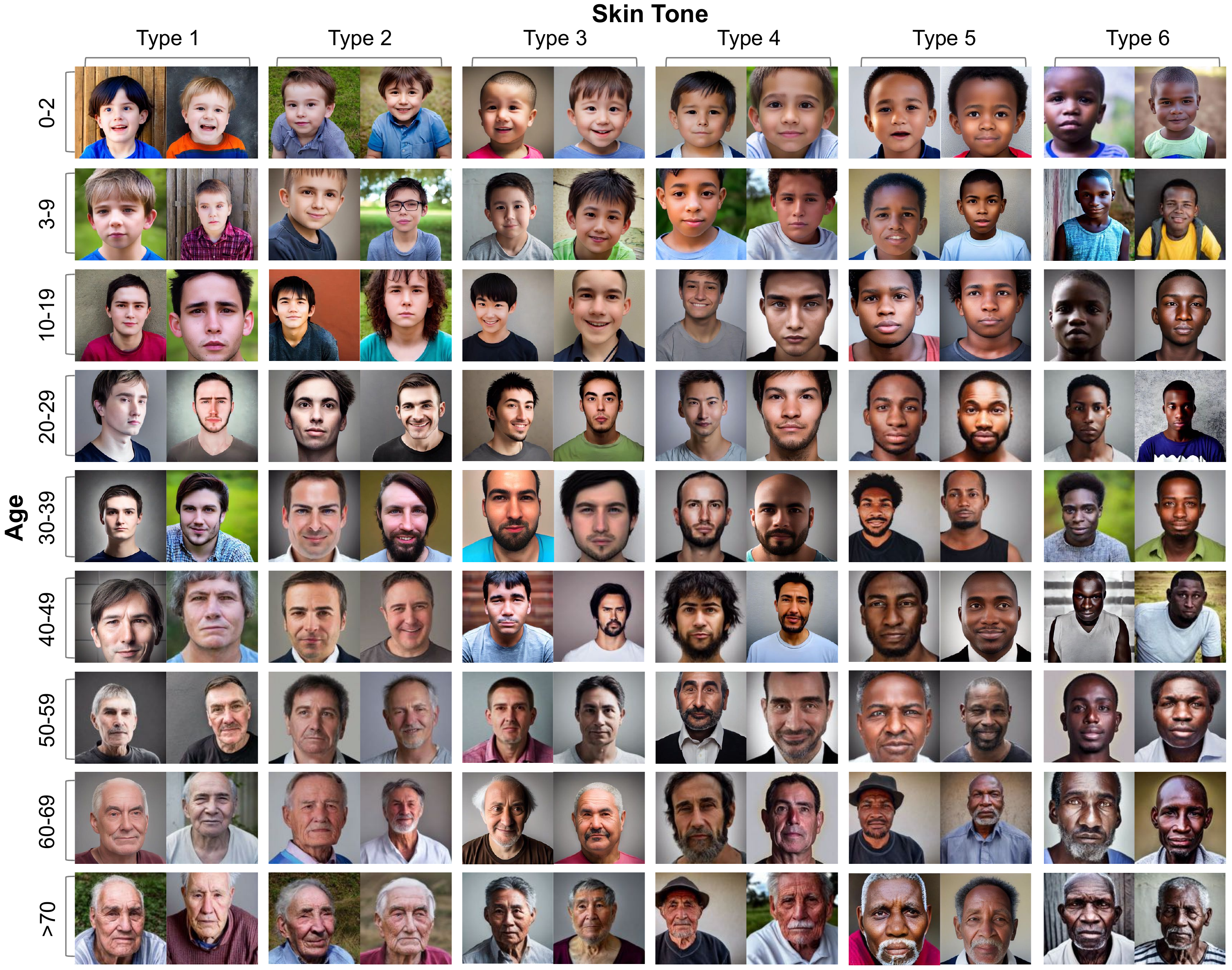}}
   \caption{\small \textbf{Results of \ourmethodbold on multi-category attributes} for Perceived Gender (2) $\times$ Skin Tone (6) $\times$ Age (9). Examples are randomly picked with ``\emph{a headshot of a person}'' for \textbf{Perceived Man} $\times$ Skin Tone (6) $\times$ Age (9). Please see Figure~\ref{fig:multi_category_woman} for more results on Perceived Woman $\times$ Skin Tone (6) $\times$ Age (9).}
    \label{fig:multi_category_man}
\end{figure*}

\begin{figure*}[t]
  \centerline{\includegraphics[width=1\linewidth]{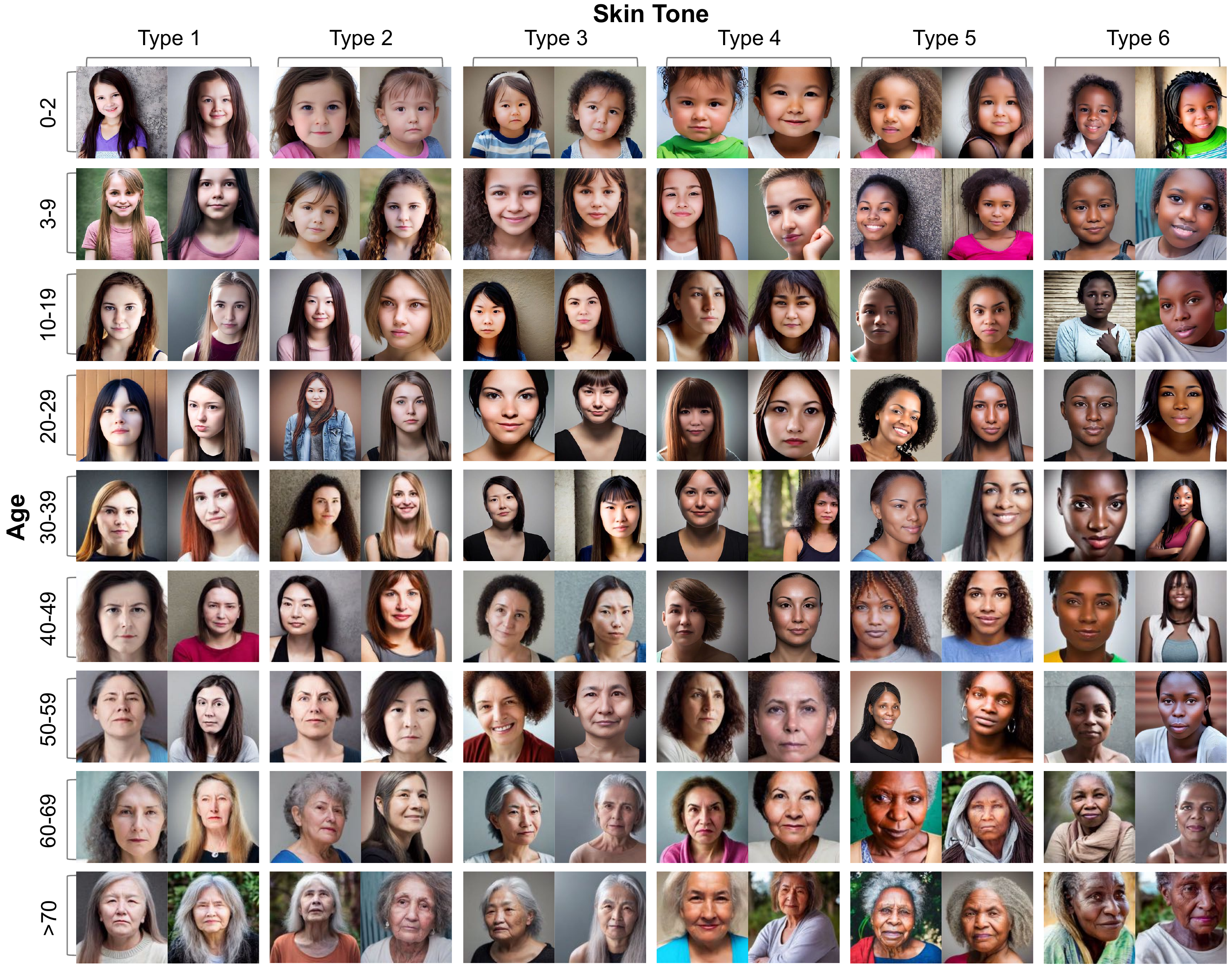}}
   \caption{\small \textbf{Results of \ourmethodbold on multi-category attributes} for Perceived Gender (2) $\times$ Skin Tone (6) $\times$ Age (9). Examples are randomly picked with ``\emph{a headshot of a person}'' for \textbf{Perceived Woman} $\times$ Skin Tone (6) $\times$ Age (9). Please see Figure~\ref{fig:multi_category_man} for more results on Perceived Man $\times$ Skin Tone (6) $\times$ Age (9).}
    \label{fig:multi_category_woman}
\end{figure*}

In Figure~\ref{fig:multi_category_hist} and Figure~\ref{fig:multi_category_qual} of the main paper, we investigated the combinations of multi-category attributes. Here, we further study another challenging setup: Perceived Gender (CelebA) $\times$ Skin Tone (FAIR) $\times$ Age (FairFace) (108 different combinations of categories in total). Qualitative results are shown in Figure~\ref{fig:multi_category_man} and in Figure~\ref{fig:multi_category_woman}. As expected, \ourmethod is capable of handling multiple fine-grained attribute categories to achieve inclusiveness.

\subsection{Other Domains}
\label{app_ss_other_domains}
\begin{figure*}[t]
  \centerline{\includegraphics[width=1\linewidth]{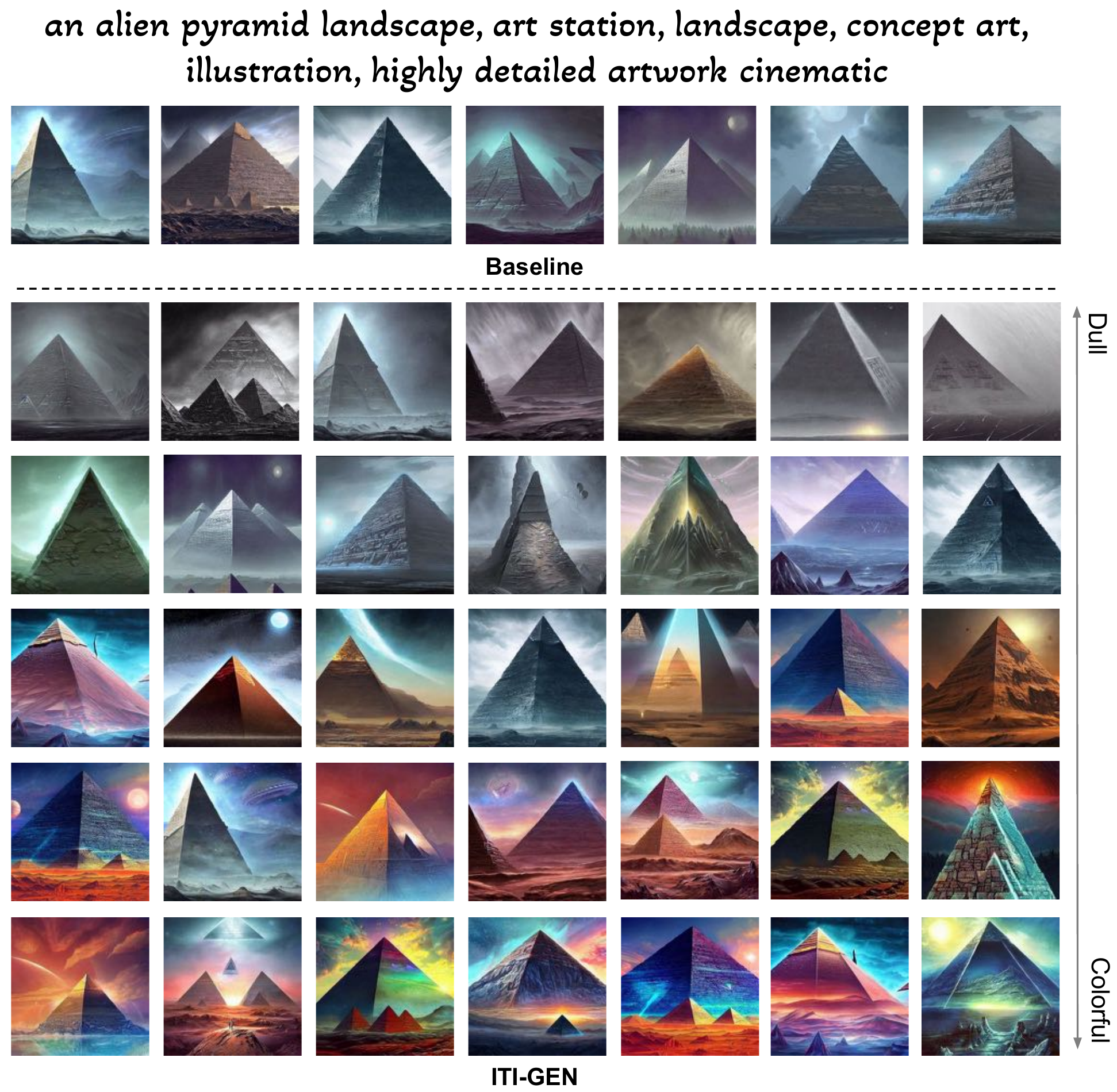}}
   \caption{\small \textbf{\ourmethodbold with perception attributes (``Colorfulness'') on scene images.}
   \ourmethod (bottom) enables the baseline Stable Diffusion (top) to generate images with different levels of colorfulness. See Section~\ref{app_s_reference_images} for details and Figure~\ref{app_fig:lhq} for reference image examples from LHQ~\cite{skorokhodov2021aligning}. Better viewed in color.}
    \label{fig:other_domain_pyramid_colorful}
\end{figure*}

\begin{figure*}[t]
  \centerline{\includegraphics[width=1\linewidth]{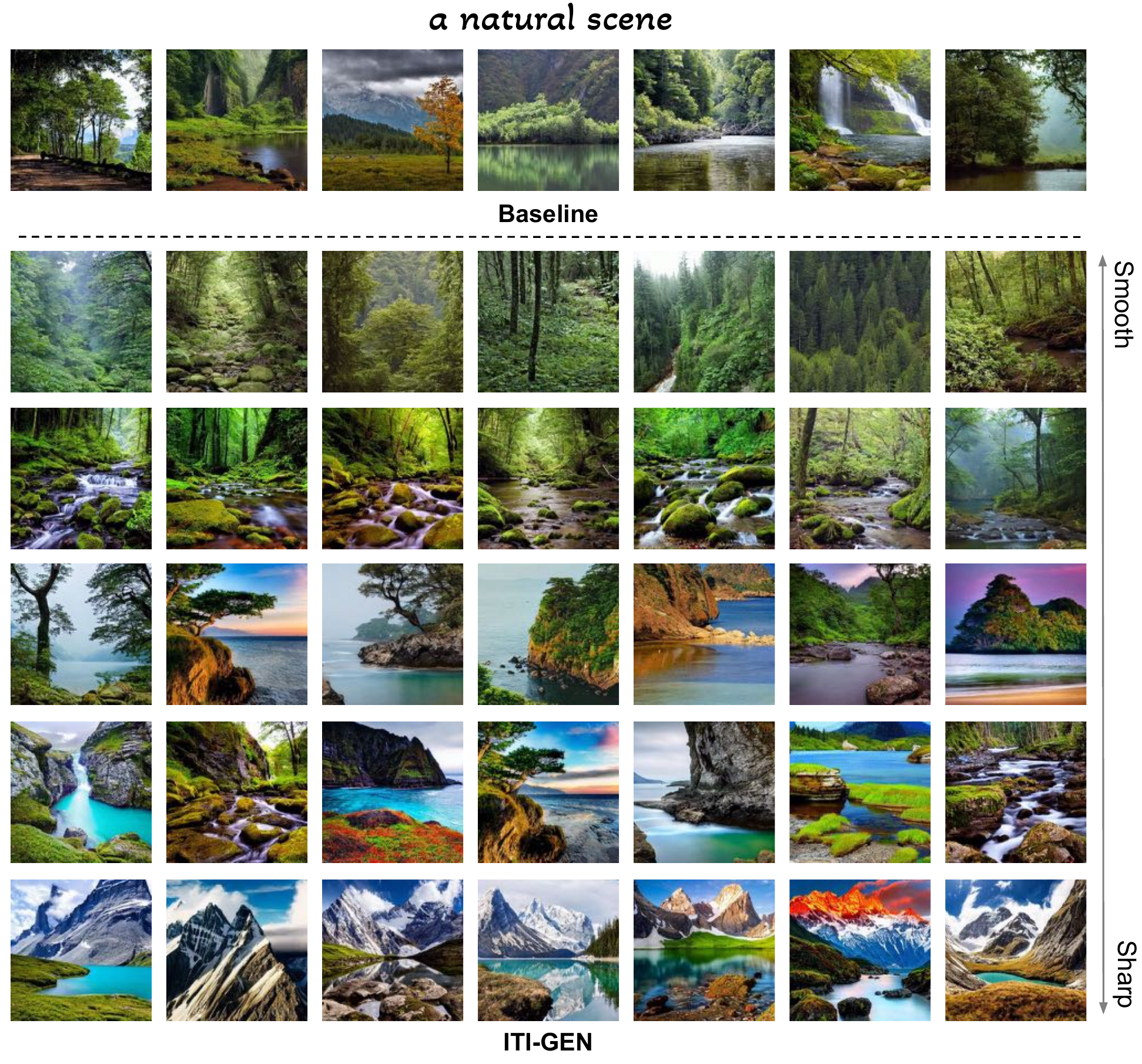}}
   \caption{\small \textbf{\ourmethodbold with perception attributes (``Sharpness'') on scene images.}
   \ourmethod (bottom) enables the baseline Stable Diffusion (top) to generate images with different levels of sharpness. See Section~\ref{app_s_reference_images} for details and Figure~\ref{app_fig:lhq} for reference image examples from LHQ~\cite{skorokhodov2021aligning}. Better viewed in color. }
    \label{fig:other_domain_natural_scene_sharp}
\end{figure*}

\begin{figure*}[t]
  \centerline{\includegraphics[width=1\linewidth]{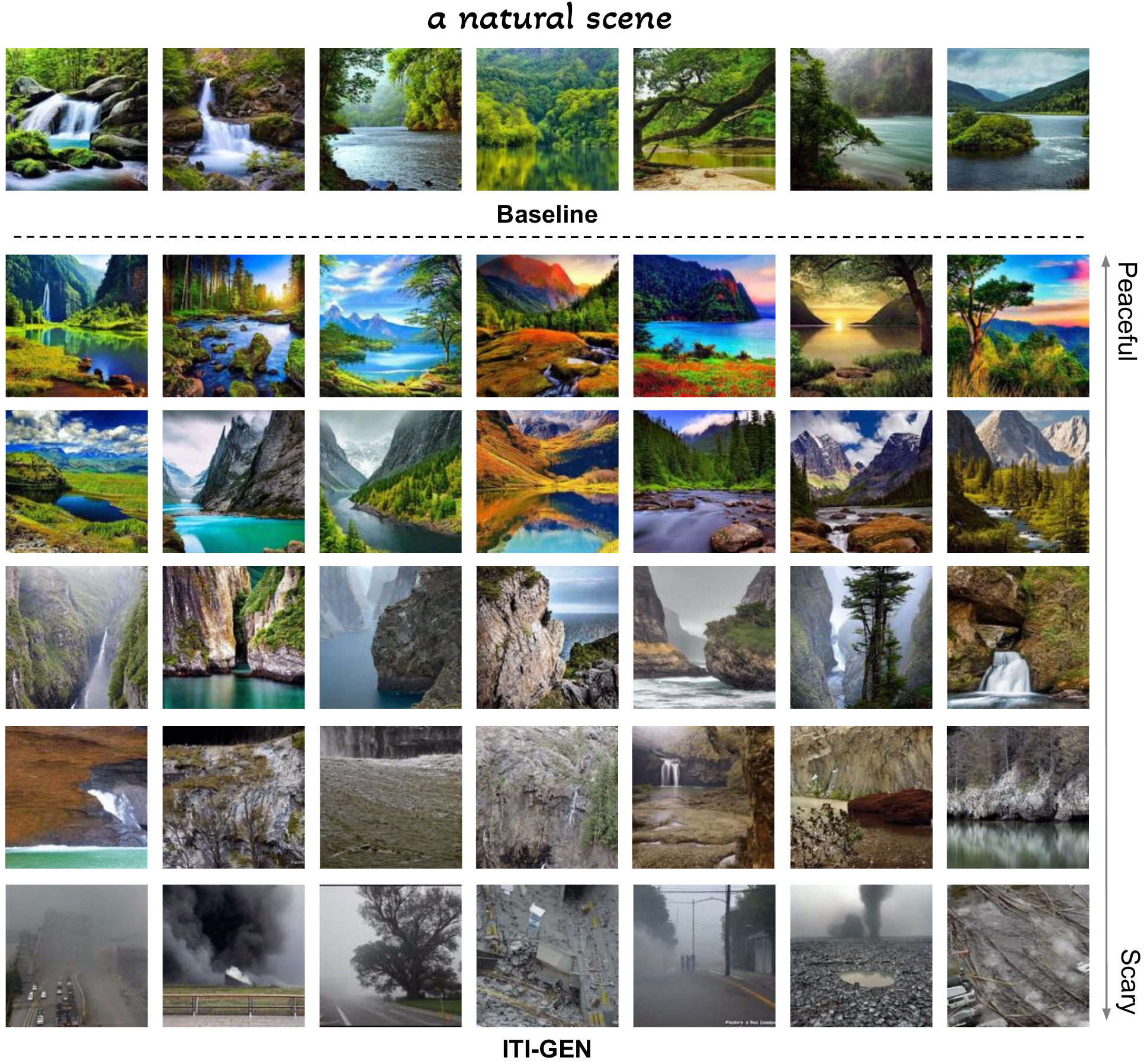}}
   \caption{\small \textbf{\ourmethodbold with perception attributes (``Scary'') on scene images.}
   \ourmethod (bottom) enables the baseline Stable Diffusion (top) to generate images with different levels of scary. See Section~\ref{app_s_reference_images} for details and Figure~\ref{app_fig:lhq} for reference image examples from LHQ~\cite{skorokhodov2021aligning}. Better viewed in color.}
    \label{fig:other_domain_natural_scene_scary}
\end{figure*}

\begin{figure*}[t]
  \centerline{\includegraphics[width=1\linewidth]{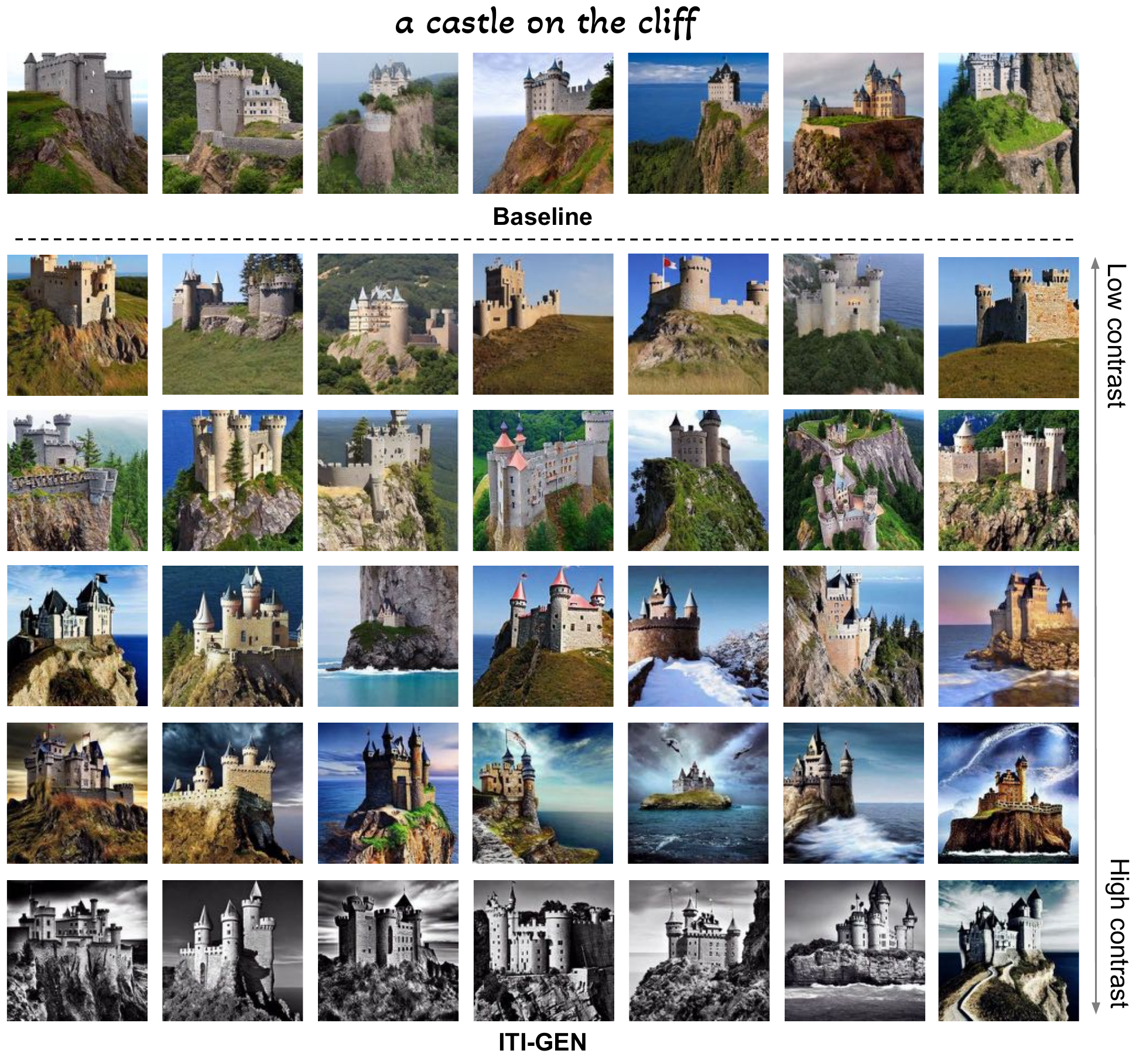}}
   \caption{\small \textbf{\ourmethodbold with perception attributes (``Contrast'') on scene images.}
   \ourmethod (bottom) enables the baseline Stable Diffusion (top) to generate images with different levels of contrast. See Section~\ref{app_s_reference_images} for details and Figure~\ref{app_fig:lhq} for reference image examples from LHQ~\cite{skorokhodov2021aligning}. Better viewed in color.}
    \label{fig:other_domain_castle_contrast}
\end{figure*}

\begin{figure*}[t]
  \centerline{\includegraphics[width=1\linewidth]{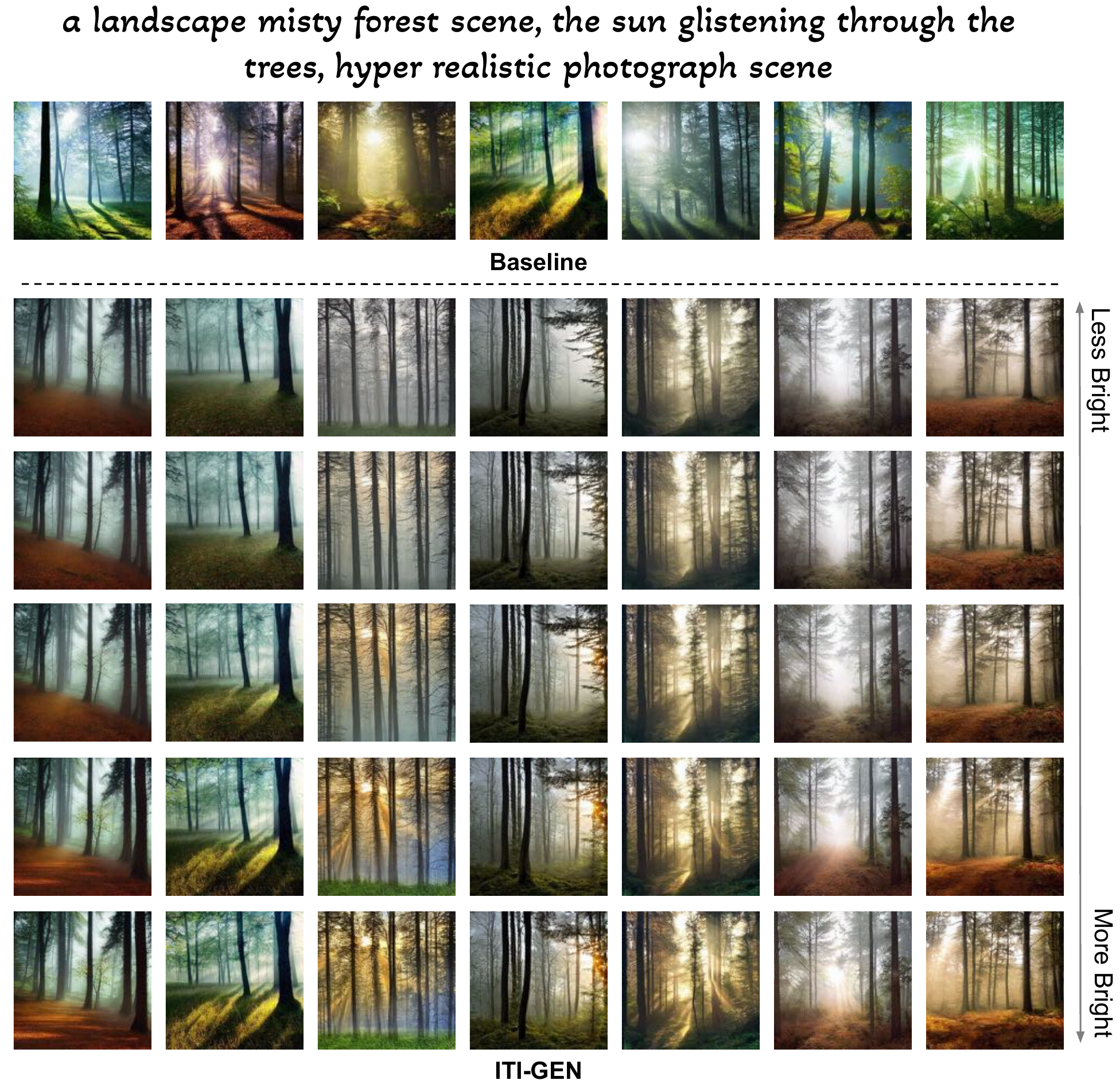}}
   \caption{\small \textbf{\ourmethodbold with perception attributes (``Brightness'') on scene images.}
   \ourmethod (bottom) enables the baseline Stable Diffusion (top) to generate images with different levels of brightness. In this example, we intentionally pick images using the same random seed in each column for \ourmethod. Please compare the first and last examples in each column for a clear change in brightness. See Section~\ref{app_s_reference_images} for details and Figure~\ref{app_fig:lhq} for reference image examples from LHQ~\cite{skorokhodov2021aligning}. Better viewed in color.}
    \label{fig:other_domain_forest_bright}
\end{figure*}

\begin{figure*}[t]
  \centerline{\includegraphics[width=1\linewidth]{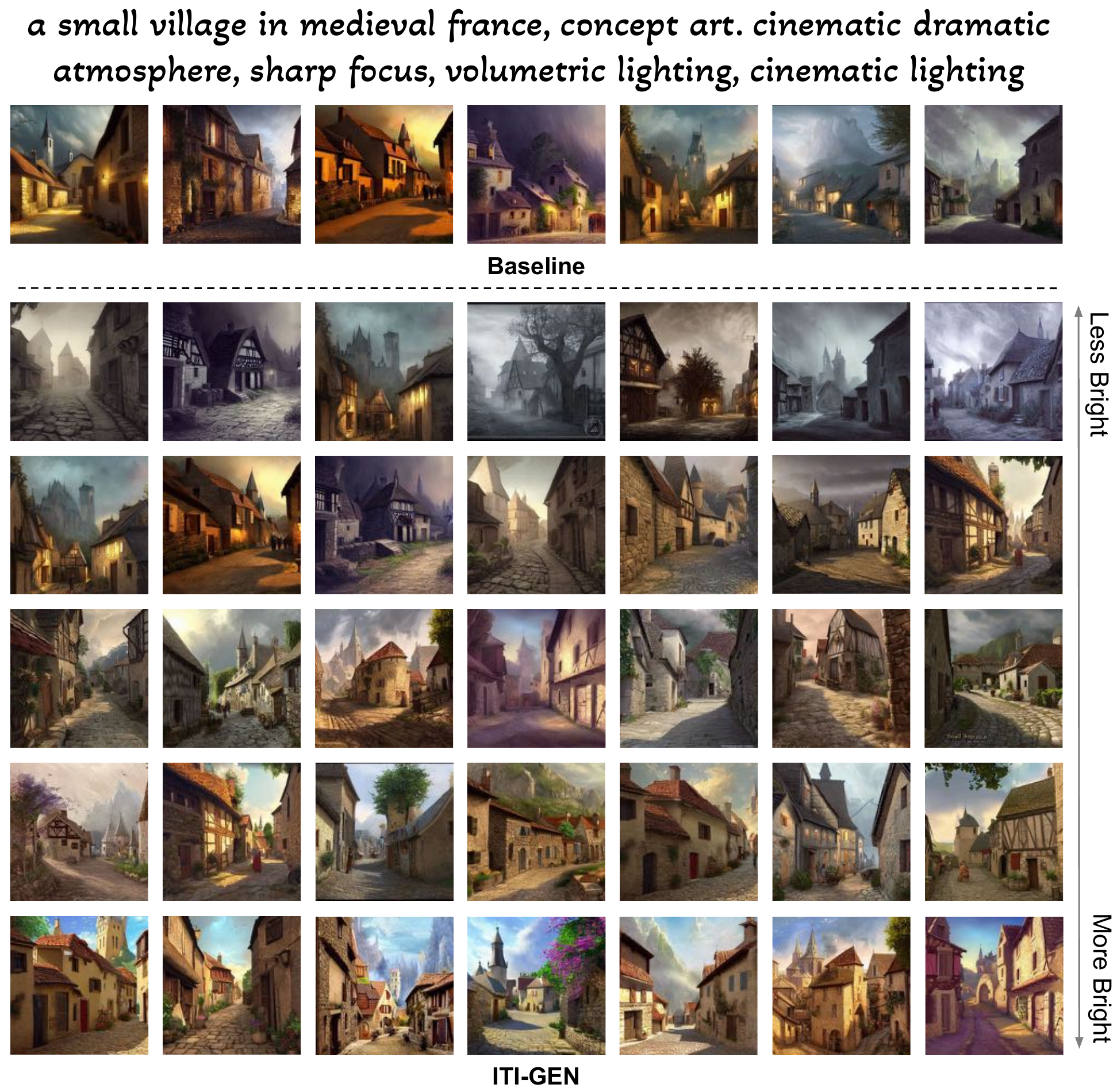}}
   \caption{\small \textbf{\ourmethodbold with perception attributes (``Brightness'') on scene images.}
   \ourmethod (bottom) enables the baseline Stable Diffusion (top) to generate images with different levels of brightness. See Section~\ref{app_s_reference_images} for details and Figure~\ref{app_fig:lhq} for reference image examples from LHQ~\cite{skorokhodov2021aligning}. Better viewed in color.}
    \label{fig:other_domain_village_bright}
\end{figure*}

As shown in Figure~\ref{fig:scene} of the main paper, \ourmethod can generalize to a different domain for perception attributes on scene images. In this subsection, we demonstrate more results of other attributes in 
Figure~\ref{fig:other_domain_pyramid_colorful} for ``colorfulness'',
Figure~\ref{fig:other_domain_natural_scene_sharp} for ``sharpness'', 
Figure~\ref{fig:other_domain_natural_scene_scary} for ``scary'',
Figure~\ref{fig:other_domain_castle_contrast} for ``contrast'',
Figure~\ref{fig:other_domain_forest_bright} for ``brightness'', and
Figure~\ref{fig:other_domain_village_bright} for ``'brightness''.
As we observe, \ourmethod generates more diverse results than the baseline model even with very complex input prompts.

\subsection{Train-once-for-all Generalization}
\label{app_ss_train_once}
\begin{figure*}[t]
  \centerline{\includegraphics[width=0.95\linewidth]{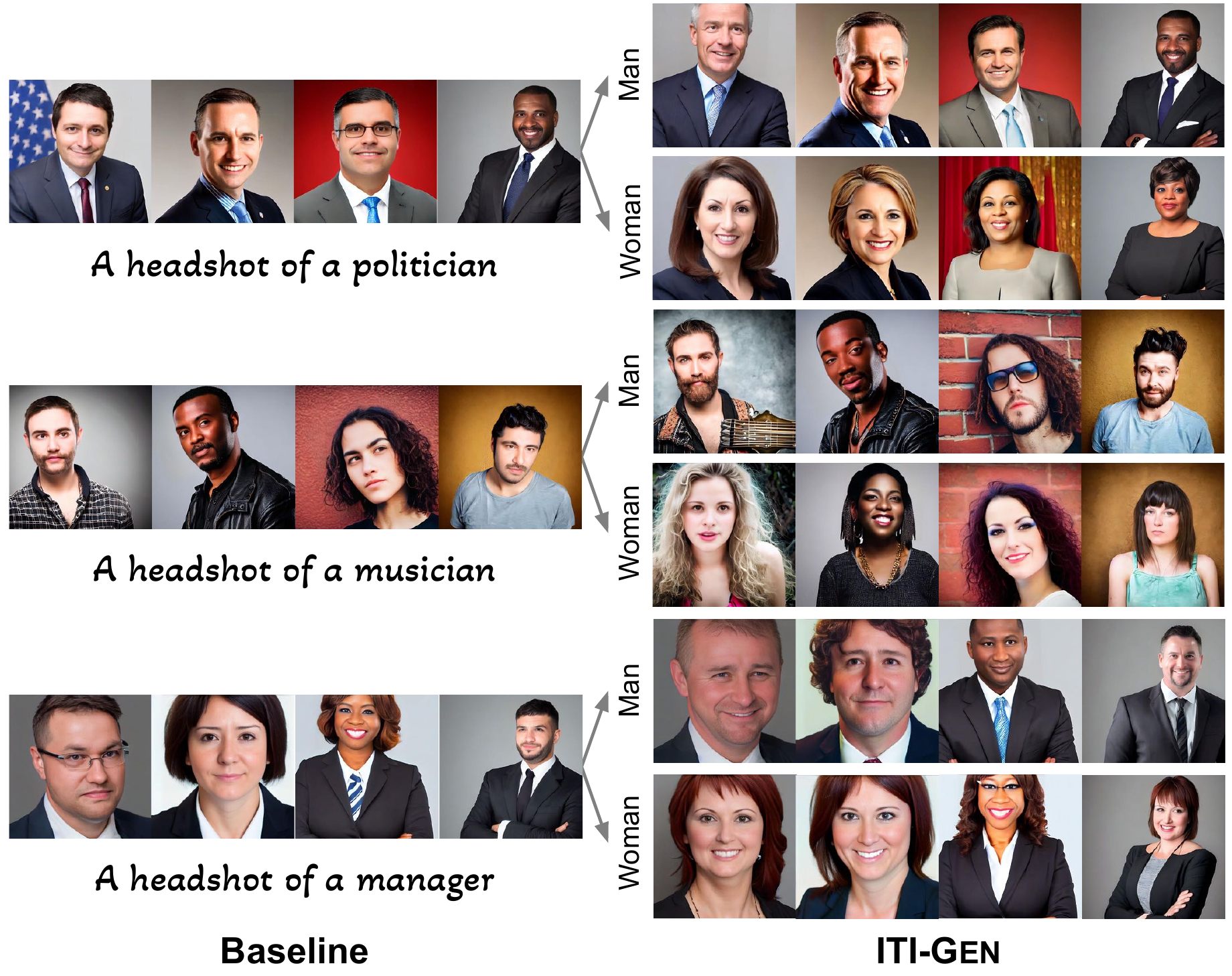}}
   \caption{\small \textbf{Additional results on train-once-for-all generalization.} Inclusive tokens of \ourmethod trained with a neutral prompt (``\emph{a headshot of a person}'') can be applied to out-of-domain prompts in these three examples to alleviate stereotypes.}
    \label{app_fig:train_once_for_all}
\end{figure*}

\begin{figure*}[t]
  \centerline{\includegraphics[width=0.95\linewidth]{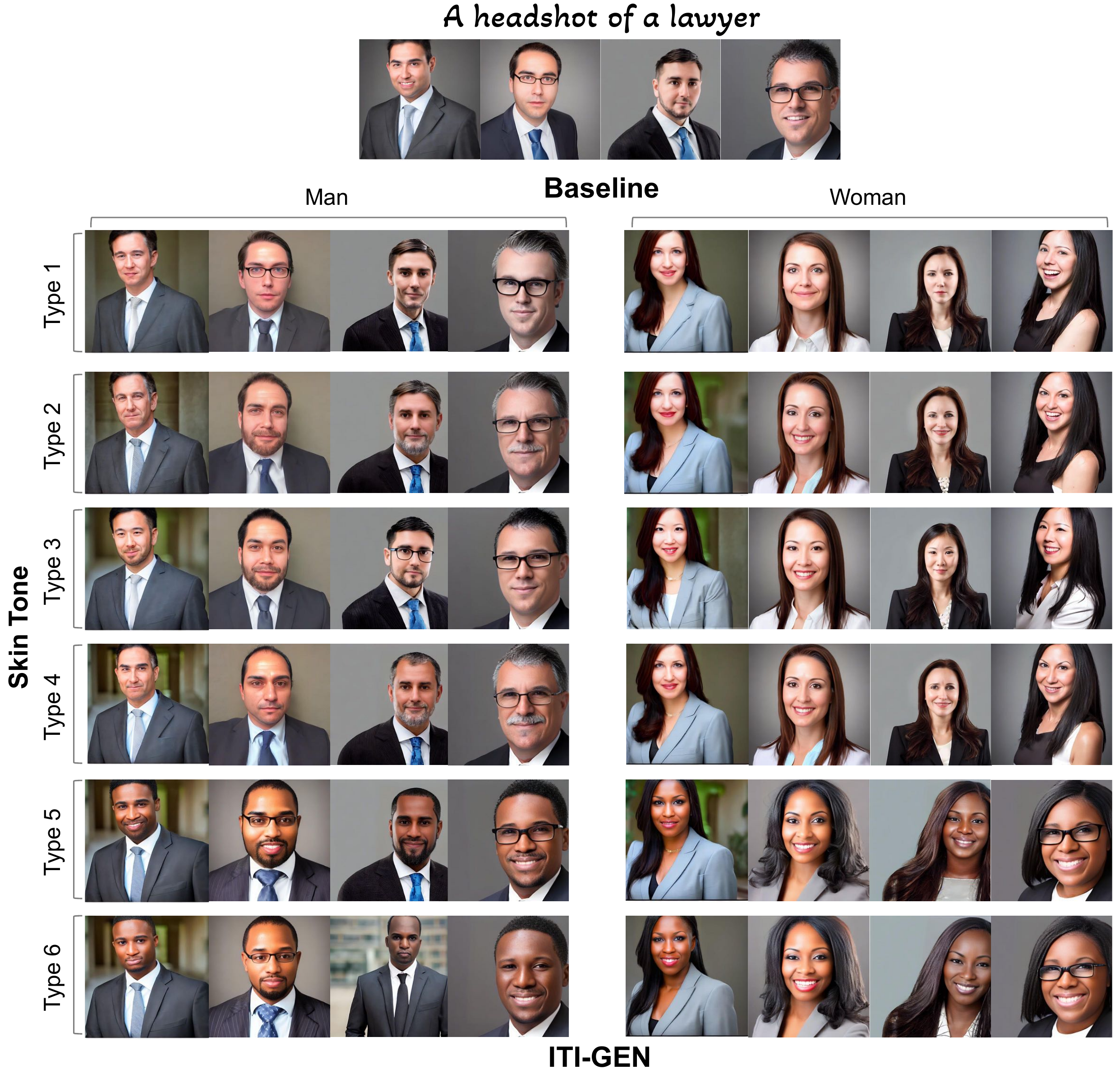}}
   \caption{\small \textbf{Additional results on train-once-for-all generalization.} Inclusive tokens of \ourmethod trained with a neutral prompt (``\emph{a headshot of a person}'') can be applied to out-of-domain prompts in these three examples to alleviate stereotypes.}
    \label{app_fig:train_once_for_all_lawyer}
\end{figure*}

\begin{figure*}[t]
  \centerline{\includegraphics[width=0.95\linewidth]{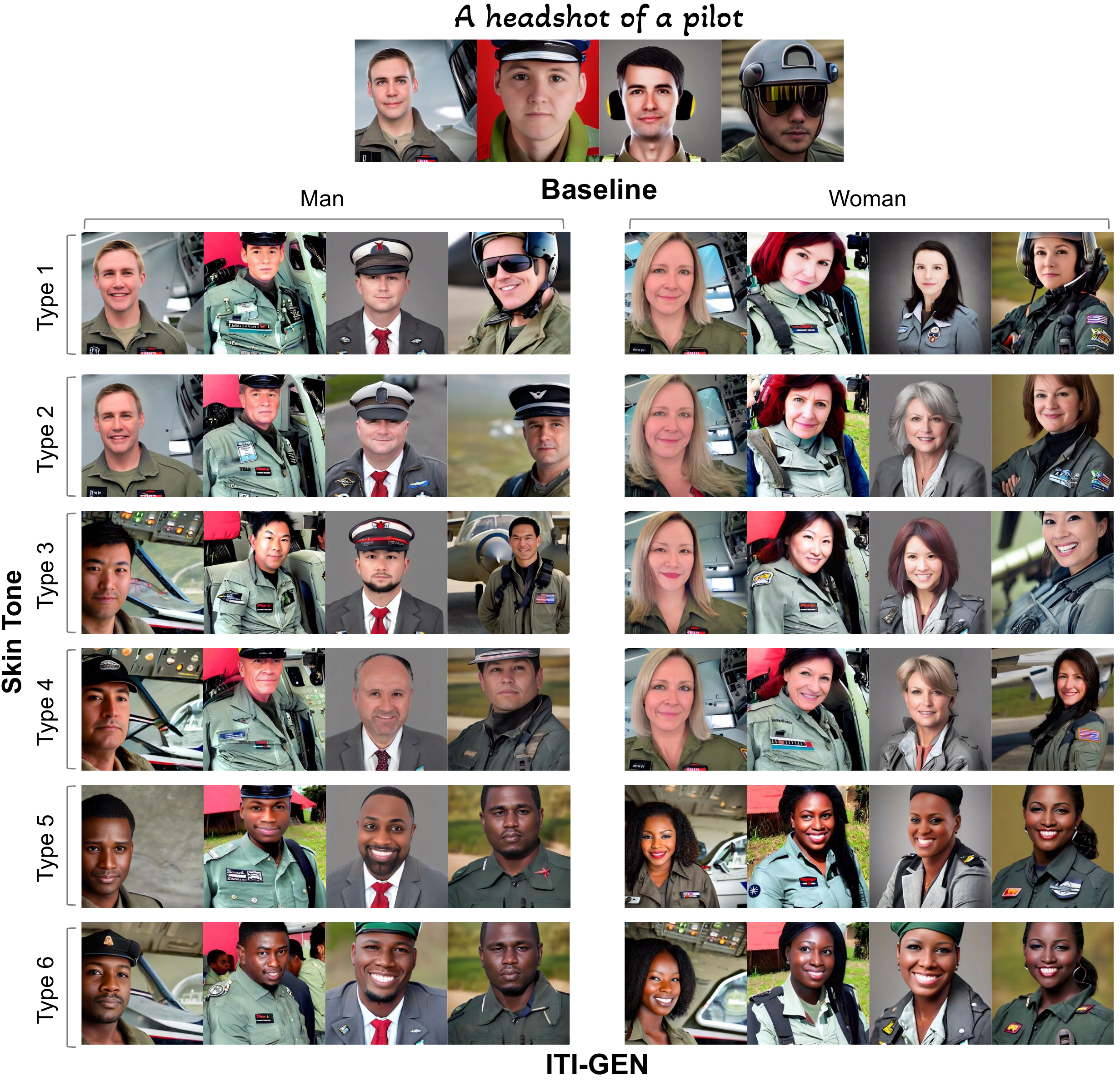}}
   \caption{\small \textbf{Additional results on train-once-for-all generalization.} Inclusive tokens of \ourmethod trained with a neutral prompt (``\emph{a headshot of a person}'') can be applied to out-of-domain prompts in these three examples to alleviate stereotypes.}
    \label{app_fig:train_once_for_all_pilot}
\end{figure*}

\begin{figure*}[t]
  \centerline{\includegraphics[width=0.95\linewidth]{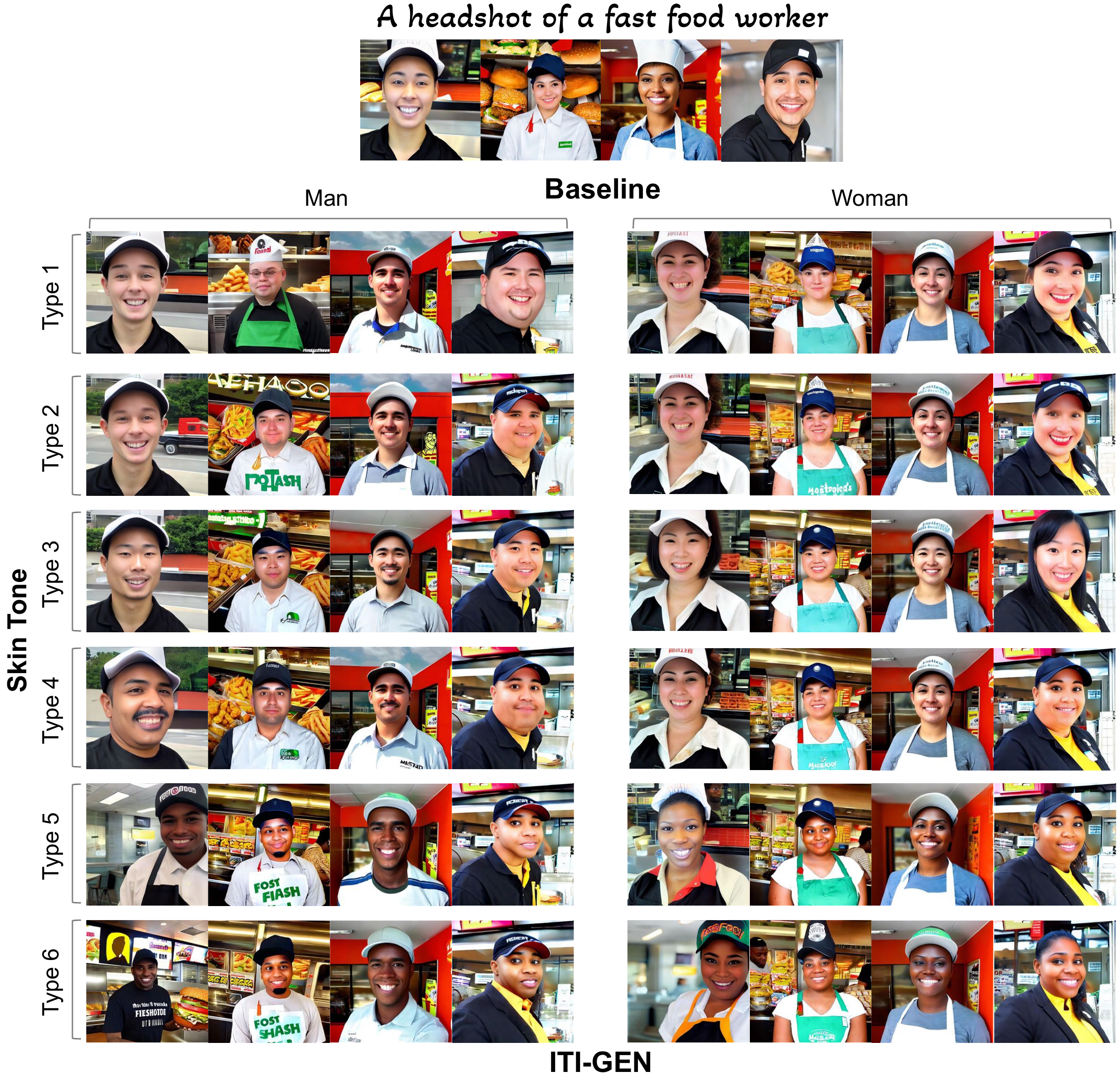}}
   \caption{\small \textbf{Additional results on train-once-for-all generalization.} Inclusive tokens of \ourmethod trained with a neutral prompt (``\emph{a headshot of a person}'') can be applied to out-of-domain prompts in these three examples to alleviate stereotypes.}
    \label{app_fig:train_once_for_all_fast_food_worker}
\end{figure*}

\begin{figure*}[t]
  \centerline{\includegraphics[width=0.95\linewidth]{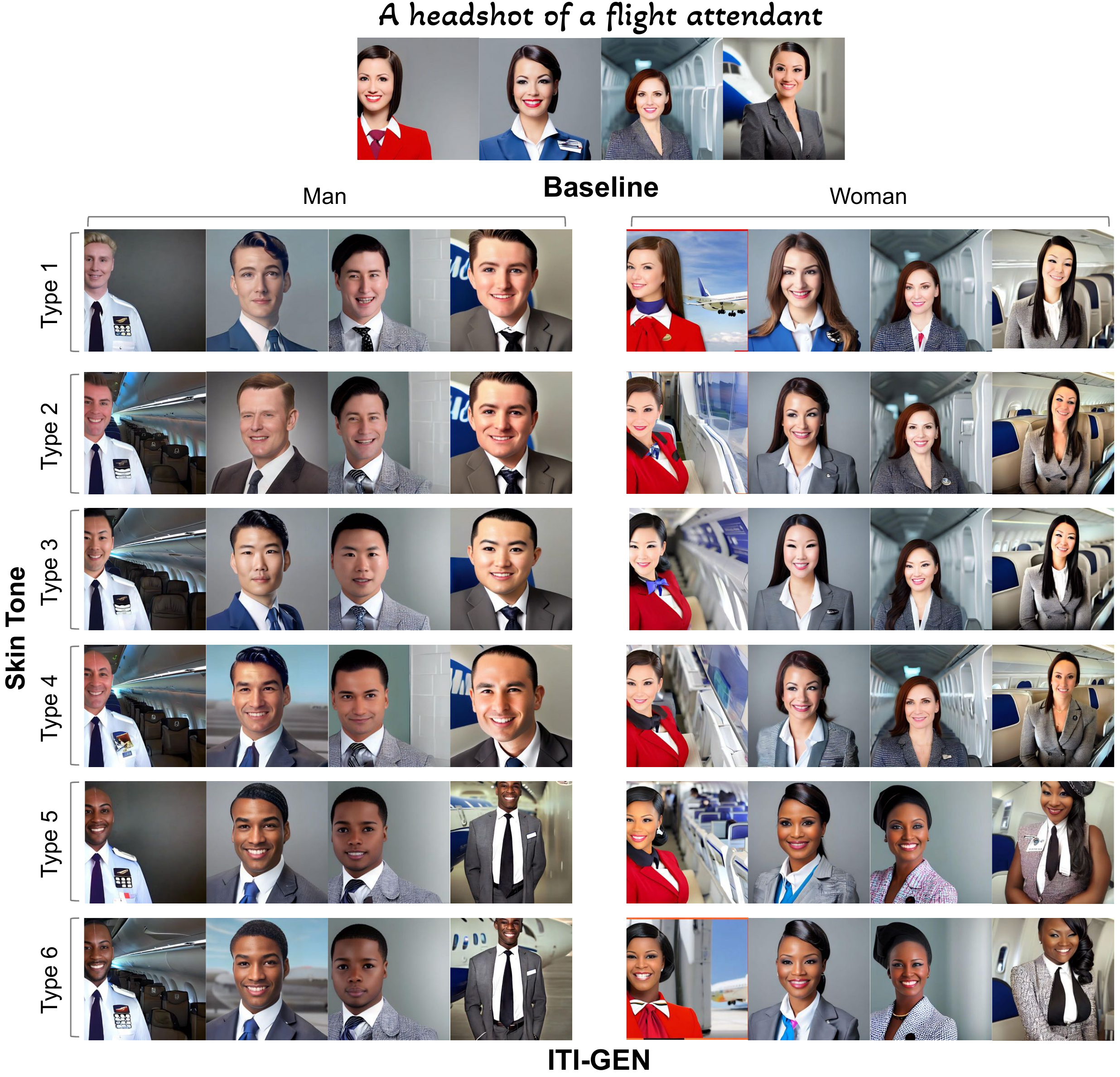}}
   \caption{\small \textbf{Additional results on train-once-for-all generalization.} Inclusive tokens of \ourmethod trained with a neutral prompt (``\emph{a headshot of a person}'') can be applied to out-of-domain prompts in these three examples to alleviate stereotypes.}
    \label{app_fig:train_once_for_all_flight}
\end{figure*}

We provide additional qualitative results with different occupation prompts in Figure~\ref{app_fig:train_once_for_all}, 
Figure~\ref{app_fig:train_once_for_all_lawyer}, 
Figure~\ref{app_fig:train_once_for_all_pilot}, 
Figure~\ref{app_fig:train_once_for_all_fast_food_worker}, and Figure~\ref{app_fig:train_once_for_all_flight}.

\subsection{Compatibility with ControlNet}
\label{app_ss_compatible}
\begin{figure*}[t]
    \centerline{\includegraphics[width=1\linewidth]{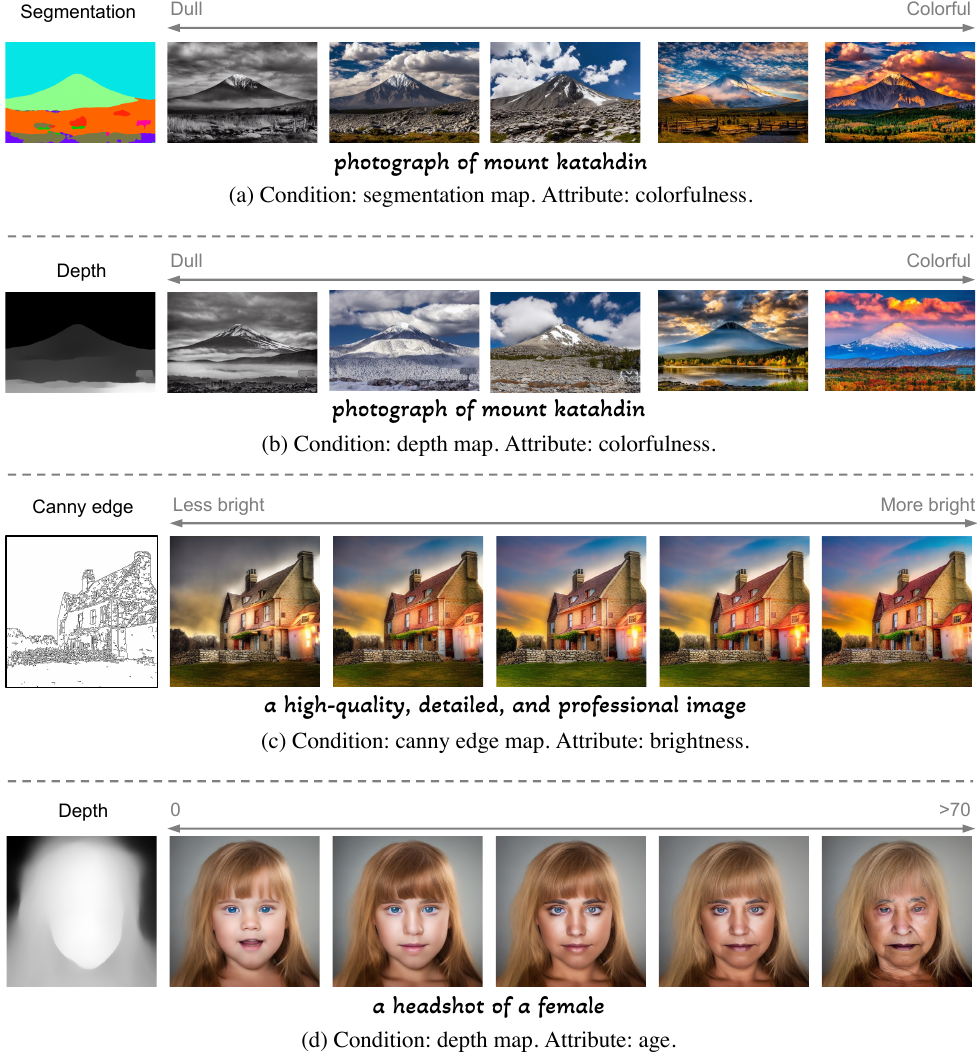}}
    \vspace{2mm}
    \caption{\small \textbf{Additional results on the compatibility with ControlNet~\cite{zhang2023adding}.} All examples are based on \emph{train-once-for-all} generation (Section~\s{3.3} of the main paper). For scene images in (a), (b), and (c), the inclusive tokens are trained with ``\emph{a natural scene}'' using LHQ images~\cite{skorokhodov2021aligning}. For human faces in (d), the tokens for age attribute are trained with ``\emph{a headshot of a person}'' using FairFace images~\cite{karkkainen2019fairface}. As discussed in Section~\ref{app_s_additional_related}, our method is designed for improving inclusiveness but not for image editing.
    }
    \label{app_fig:controlnet}
\end{figure*}

We provide additional examples of compatibility with ControlNet in Figure~\ref{app_fig:controlnet}.

\section{Future Work}
\label{app_s_future_work}
To establish the new direction and demonstrate its feasibility so that future works can easily build upon, we intentionally avoid sophisticated techniques to improve \ourmethod in favor of simplicity and believe that additional modifications can further enhance the inclusive generative models.

\mypara{Lifelong \ourmethodbold.} 
In this study, we assume all the attributes are accessible at the same time. In practice, we hope to show that \ourmethod is capable of the continue learning setup. That is, adding new attributes while without forgetting or re-training the previous inclusive tokens.

\mypara{Other Attributes.}
There are other attributes \ourmethod might be able to control via appropriately prepared reference images. For example, the 3D geometry attributes such as head poses and materials such as normal and lighting.

\end{document}